\documentclass{ijac}
\usepackage{multicol}
\usepackage{subfigure}
\usepackage{amsmath}
\usepackage{amssymb}
\usepackage{amsfonts}
\usepackage{graphicx}
\usepackage{url}
\usepackage{ccaption}
\usepackage{booktabs} 

\usepackage{times}
\usepackage{epsfig}
\usepackage{caption}
\usepackage{verbatim}
\usepackage{array}
\usepackage{xcolor}
\usepackage{multirow}
\usepackage{capt-of,etoolbox}

\firstheadname{} 
\firstfootname{} 
\headevenname{} 
\headoddname{}

\makeatletter
\apptocmd\@maketitle{{\myfigure{}\par}}{}{}
\makeatother

\begin{document}
	
	\title{Deep Audio-Visual Learning: A Survey}
	
	\author{
		Hao Zhu$^{1,2}$ \qquad Mandi Luo$^{2,3}$ \qquad  Rui Wang$^{1,2}$ \qquad  Aihua Zheng$^{1}$ \qquad  Ran He$^{2,3,4}$}
	
	\address{
		$^1$School of Computer Science and Technology, Anhui University, Hefei, China \\
		$^2$Center for Research on Intelligent Perception and Computing (CRIPAC) and National Laboratory of Pattern Recognition (NLPR), CASIA, Beijing, China \\
		$^3$School of Artificial Intelligence, University of the Chinese Academy of Sciences, Beijing, China \\
		$^4$Center for Excellence in Brain Science and Intelligence Technology, CAS, Beijing, China
	}
	
\abstract{
	Audio-visual learning, aimed at exploiting the relationship between audio and visual modalities, has drawn considerable attention since deep learning started to be used successfully. Researchers tend to leverage these two modalities either to improve the performance of previously considered single-modality tasks or to address new challenging problems. In this paper, we provide a comprehensive survey of recent audio-visual learning development.
	We divide the current audio-visual learning tasks into four different subfields: audio-visual separation and localization, audio-visual correspondence learning, audio-visual generation, and audio-visual representation learning. State-of-the-art methods as well as the remaining challenges of each subfield are further discussed. Finally, we summarize the commonly used datasets and performance metrics.
	}
	

	\newcommand\myfigure{%
		\centering
		\includegraphics[width=0.8\linewidth]{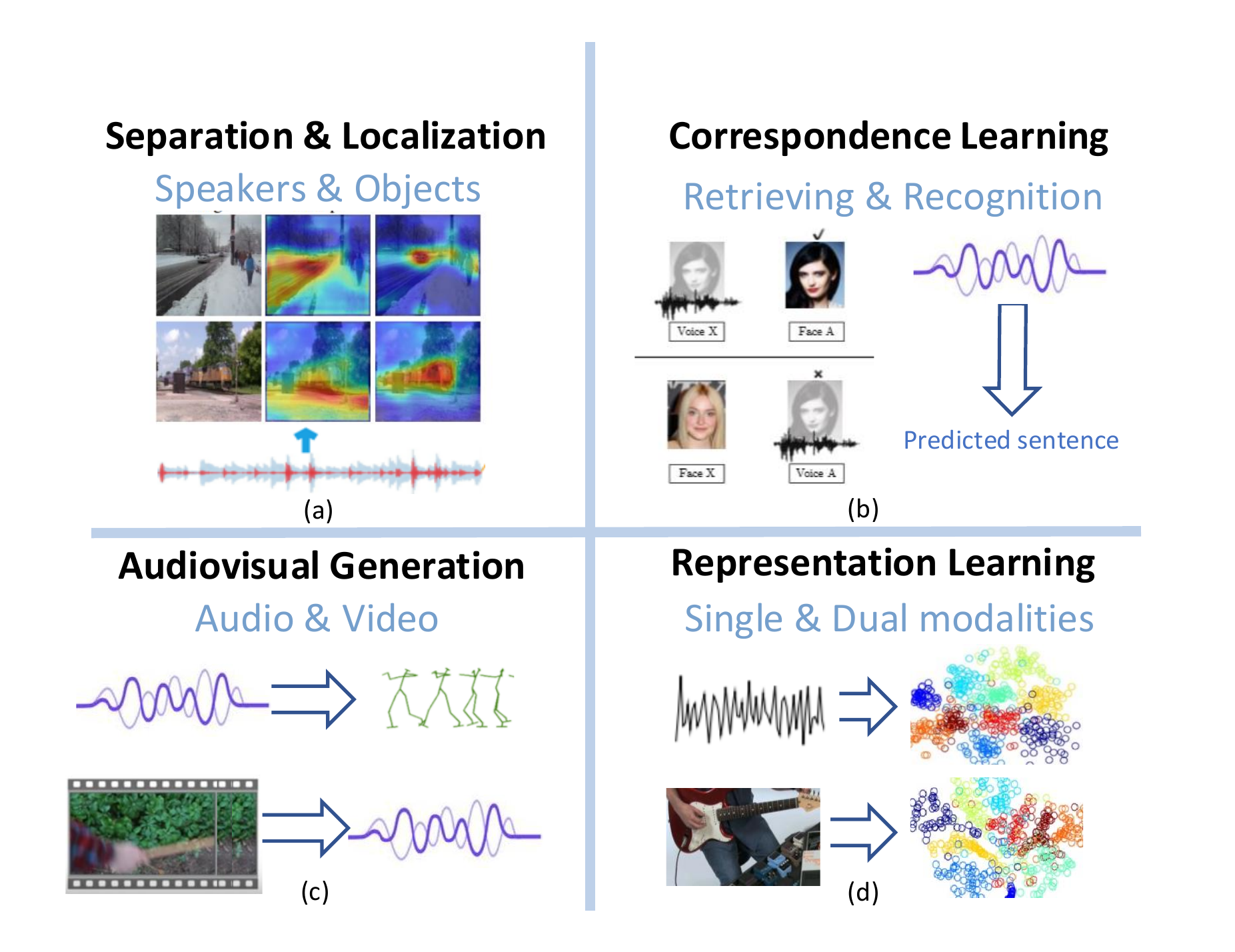}
		\captionof{figure}{Illustration of four categories of tasks in audio-visual learning.}
		\label{fig:overall}
	}
		
	\maketitle
	
	\pagestyle{ijacheadings}
		
	\section{Introduction}

    Human perception is multidimensional and includes vision, hearing, touch, taste, and smell. In recent years, along with the vigorous development of artificial intelligence technology, the trend from single-modality learning  to multimodality learning has become crucial to better machine perception. 
    Analyses of audio and visual information, representing the two most important perceptual modalities in our daily life, have been widely developed in both academia and industry in the past decades. Prominent achievements include speech recognition~\cite{shannon1995speech, krishna2019speech}, facial recognition~\cite{he2010maximum, fu2019dual}, etc. Audio-visual learning (AVL) using both modalities has been introduced to overcome the limitation of perception tasks in each modality. In addition, exploring the relationship between audio and visual information leads to more interesting and important research topics and ultimately better perspectives on machine learning.

	The purpose of this article is to provide an overview of the key methodologies in audio-visual learning, which aims to discover the relationship between audio and visual data for many challenging tasks. In this paper, we mainly divide these efforts into four categories: (1) audio-visual separation and localization, (2) audio-visual corresponding learning, (3) audio and visual generation, and (4) audio-visual representation.

	
	
    {\bf Audio-visual separation and localization} aim to separate specific sounds emanating from the corresponding objects and localize each sound in the visual context, as illustrated in Fig. \ref{fig:overall} (a). 
    Audio separation has been investigated extensively in the signal processing community during the past two decades. With the addition of the visual modality, audio separation can be transformed into audio-visual separation, which has proven to be more effective in noisy scenes \cite{gabbay2018seeing, afouras2018conversation, ephrat2018looking}. Furthermore, introducing the visual modality allows for audio localization, i.e., the localization of a sound in the visual modality according to the audio input.
    The tasks of audio-visual separation and localization themselves not only lead to valuable applications but also provide the foundation for other audio-visual tasks, e.g., generating spatial audio for 360$^\circ$ video \cite{morgado2018self}. 
    Most studies in this area focus on unsupervised learning due to the lack of training labels.

    
    {\bf Audio-visual correspondence learning} focuses on discovering the global semantic relation between audio and visual modalities, as shown in Fig. \ref{fig:overall} (b). It consists of audio-visual retrieval and audio-visual speech recognition tasks. 
    The former uses audio or an image to search for its counterpart in another modality, while the latter derives from the conventional speech recognition task that leverages visual information to provide a more semantic prior to improve recognition performance. 
    Although both of these two tasks have been extensively studied, they still entail major challenges, especially for fine-grained cross-modality retrieval and homonyms in speech recognition.


    {\bf Audio-visual generation} tries to synthesize 
the other modality based on one of them, which is different from the above two tasks leveraging both audio and visual modalities as inputs.
    Trying to make a machine that is creative is always challenging, and many generative models have been proposed \cite{gulrajani2017improved, karras2019style}. 
    Audio-visual cross-modality generation has recently drawn considerable attention. It aims to generate audio from visual signals, or vice versa. 
    Although it is easy for a human to perceive the natural correlation between sounds and appearance, this task is challenging for machines due to heterogeneity across modalities. 
    As shown in Fig. \ref{fig:overall} (c), vision to audio generation mainly focuses on recovering speech from lip sequences or predicting the sounds that may occur in the given scenes. In contrast, audio to vision generation can be classified into three categories: audio-driven image generation, body motion generation, and talking face generation.


    The last task---\textbf{audio-visual representation learning}---aims to automatically discover the representation from raw data. 
    A human can easily recognize audio or video based on long-term brain cognition. 
    However, machine learning algorithms such as deep learning models are heavily dependent on data representation. Therefore, learning suitable data representations for machine learning algorithms may improve performance. 

    Unfortunately, real-world data such as images, videos and audio do not possess specific algorithmically defined features \cite{bengio2013representation}.
    Therefore, an effective representation of data determines the success of machine learning algorithms. Recent studies seeking better representation have designed various tasks, such as audio-visual correspondence (AVC) \cite{arandjelovic2017look} and audio-visual temporal synchronization (AVTS) \cite{korbar2018co}. By leveraging such a learned representation, one can more easily solve audio-visual tasks mentioned in the very beginning.

    In this paper, we present a comprehensive survey of the above four directions of audio-visual learning. The rest of this paper is organized as follows. We introduce the four directions in Secs. \ref{sec:sep_loc}, \ref{sec:corr}, \ref{sec:generation} and \ref{sec:representation_learning}. Sec. \ref{sec:dataset} summarizes the commonly used public audio-visual datasets. Finally, Sec. \ref{sec:conclusion} concludes the paper.

	\section{Audio-visual Separation and Localization}
	\label{sec:sep_loc}
	
\begin{figure}[tb]
		\includegraphics[width=1\linewidth]{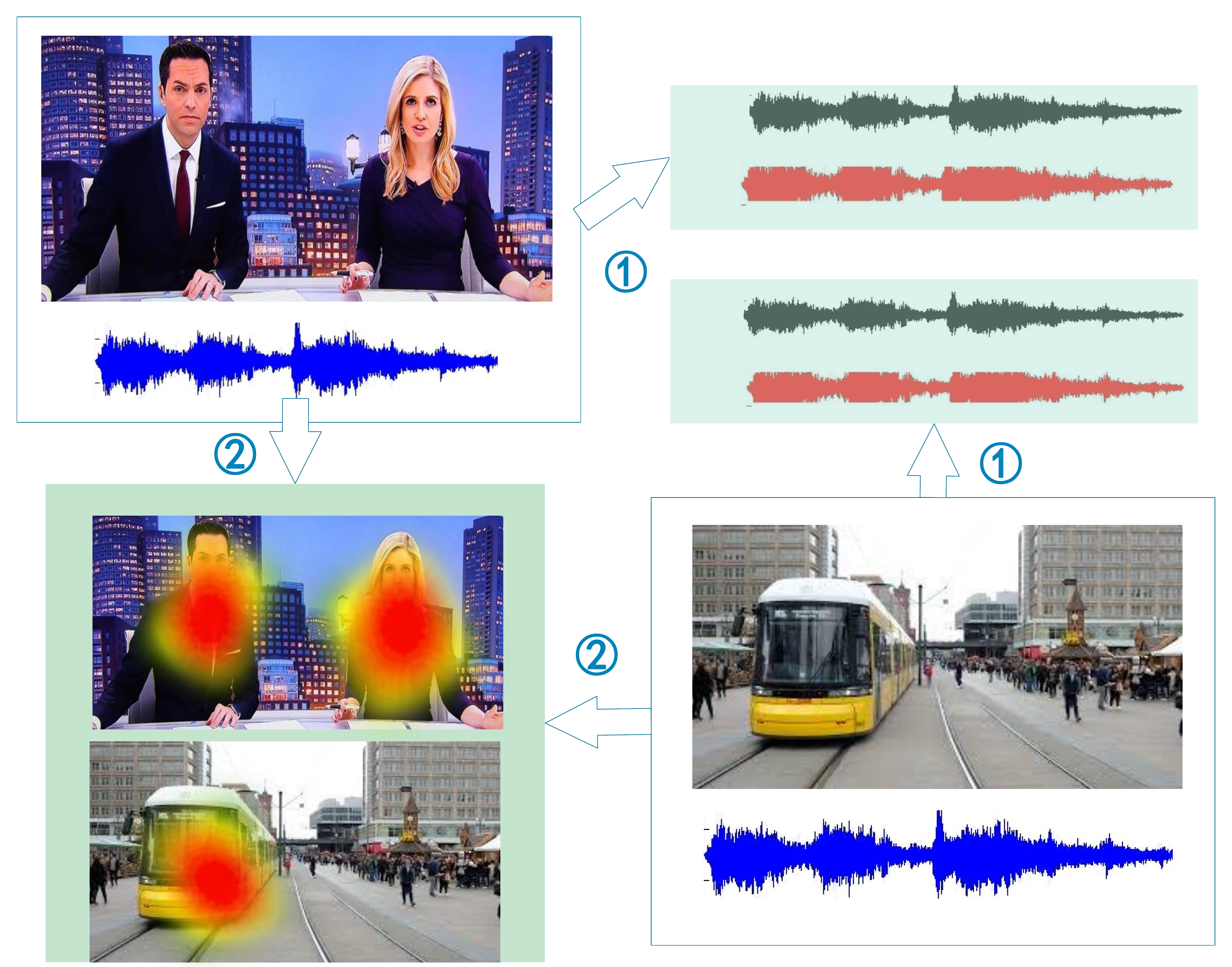}
		\caption{Illustration of audio-visual separation and localization task. Paths 1 and 2 denote separation and localization tasks, respectively.} 
		\label{fig:localization-intro} 
	\end{figure}

    The objective of audio-visual separation is to separate different sounds from the corresponding objects, while audio-visual localization mainly focuses on localizing a sound in a visual context.
    As shown in Fig. \ref{fig:localization-intro}, we classify types of this task by different identities: speakers (Fig. \ref{fig:localization-intro} (a)) and objects (Fig. \ref{fig:localization-intro} (b)).
    The former concentrates on a person's speech that can be used for television programs to enhance the target speakers' voice, while the latter is a more general and challenging task that separates arbitrary objects rather than speakers only.
    In this section, we provide an overview of these two tasks, examining the motivations, network architectures, advantages, and disadvantages.

	\subsection{Speaker Separation}

    The speaker separation task is a challenging task and is also known as the `cocktail party problem'. It aims to isolate a single speech signal in a noisy scene. 
    Some studies tried to solve the problem of audio separation with only the audio modality and achieved exciting results \cite{isik2016single,luo2018speaker}. Advanced approaches~\cite{gabbay2018seeing, ephrat2018looking} tried to utilize visual information to aid the speaker separation task and significantly surpassed single modality-based methods. 
    The early attempts leveraged mutual information to learn the joint distribution between the audio and the video \cite{darrell2000audio, fisher2001audiovisualfusion}.
    Subsequently, several methods focused on analyzing videos containing salient motion signals and the corresponding audio events (e.g., a mouth starting to move or a hand on piano suddenly accelerating) \cite{bochen2017see, pu2017audio}.
    
	\begin{figure}[h]
		\includegraphics[width=0.9\linewidth]{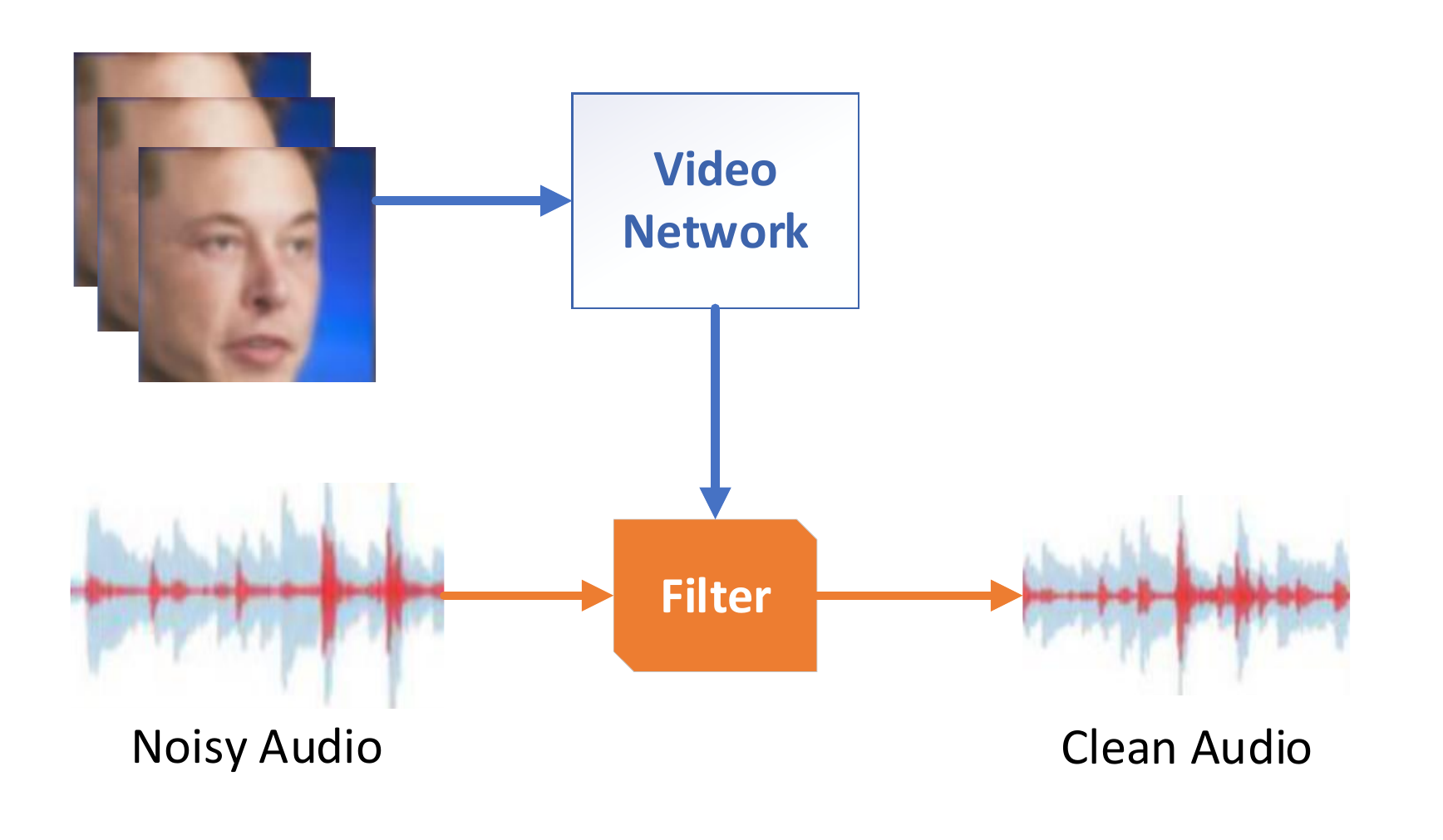}
		\caption{Basic pipeline of a noisy audio filter.}
		\label{fig:audiofilter}
	\end{figure}

    Gabbay et al.\cite{gabbay2018seeing} proposed isolating the voice of a specific speaker and eliminating other sounds in an audio-visual manner. 
    Instead of directly extracting the target speaker's voice from the noisy sound, which may bias the training model, the researchers first fed the video frames into a video-to-speech model and then predicted the speaker's voice by the facial movements captured in the video. Afterwards, the predicted voice was used to filter the mixtures of sounds, as shown in Fig. \ref{fig:audiofilter}.
    
    Although Gabbay et al. \cite{gabbay2018seeing} improved the quality of separated voice by adding the visual modality, their approach was only applicable in controlled environments. 
    To obtain intelligible speech in an unconstrained environment, Afouras et al. \cite{afouras2018conversation} proposed a deep audio-visual speech enhancement network to separate the speaker's voice of the given lip region by predicting both the magnitude and phase of the target signal. The authors treated the spectrograms as temporal signals rather than images for a network. Additionally, instead of directly predicting clean signal magnitudes, they also tried to generate a more effective soft mask for filtering.

    In contrast to previous approaches that require training a separate model for each speaker of interest (speaker-dependent models), Ephrat et al. \cite{ephrat2018looking} proposed a speaker-independent model that was only trained once and was then applicable to any speaker. This approach even outperformed the state-of-the-art speaker-dependent audio-visual speech separation methods.
    The relevant model consists of multiple visual streams and one audio stream, concatenating the features from different streams into a joint audio-visual representation. This feature is further processed by a bidirectional LSTM and three fully connected layers. Finally, an elaborate spectrogram mask is learned for each speaker to be multiplied by the noisy input. Finally, the researchers converted it back to waveforms to obtain an isolated speech signal for each speaker. 
    Lu et al. \cite{lu2018listen} designed a network similar to that of \cite{ephrat2018looking}. The difference is that the authors enforced an audio-visual matching network to distinguish the correspondence between speech and human lip movements. Therefore, they could obtain clear speech. 
	
    Instead of directly utilizing video as a condition, Morrone et al. \cite{morrone2019face} further introduced landmarks as a fine-grained feature to generate time-frequency masks to filter mixed-speech spectrogram. 
    
    \subsection{Separating and Localizing Objects' Sounds}
    Instead of matching a specific lip movement from a noisy environment as in the speaker separation task, humans focus more on objects while dealing with sound separation and localization. It is difficult to find a clear correspondence between audio and visual modalities due to the challenge of exploring the prior sounds from different objects. 
	
	\subsubsection{Separation}
    The early attempt to solve this localization problem can be traced back to 2000 \cite{Hershey00audio-vision} and a study that synchronized low-level features of sounds and videos. Fisher et al. \cite{fisher2001audiovisualfusion} later proposed using a nonparametric approach to learn a joint distribution of visual and audio signals and then project both of them to a learned subspace.  
    Furthermore, several acoustics-based methods \cite{van2004optimum, zunino2015seeing} were described that required specific devices for surveillance and instrument engineering, such as microphone arrays used to capture the differences in the arrival of sounds.
    
    
    To learn audio source separation from large-scale in-the-wild videos containing multiple audio sources per video, Gao et al. \cite{gao2018object-sounds} suggested learning an audio-visual localization model from unlabeled videos and then exploiting the visual context for audio source separation.
    Researchers' approach relied on a multi-instance multilabel learning framework to disentangle the audio frequencies related to individual visual objects even without observing or hearing them in isolation. The multilabel learning framework was fed by a bag of audio basis vectors for each video, and then, the bag-level prediction of the objects presented in the audio was obtained.

	\begin{table*}[h]
		\centering
		\label{tab:localization_separation_summary}
		\caption{Summary of recent audio-visual separation and localization approaches.}	\begin{tabular}{ccll} 
			\hline
			\textbf{Category}  & Method & Ideas \& Strengths & Weaknesses \\ 
			\hline
			& Gabbay et al. \cite{gabbay2018seeing} & \begin{tabular}[l]{@{}l@{}}Predict speaker's voice based on\\faces in video used as a filter \end{tabular} & \begin{tabular}[l]{@{}l@{}}Can only be used \\ in controlled environments \end{tabular} \\ 
			\cline{2-4}
			& Afouras et al. \cite{afouras2018conversation} & \begin{tabular}[l]{@{}l@{}}Generate a soft mask for \\filtering in the wild\end{tabular} & \begin{tabular}[l]{@{}l@{}}Requires training a \\ separate model for \\ each speaker of interest \end{tabular} \\ 
			\cline{2-4}
			
			Speaker Separation & Lu et al. \cite{lu2018listen} & \begin{tabular}[l]{@{}l@{}}Distinguish the correspondence \\ between speech and human \\ speech lip movements \end{tabular} & \multicolumn{1}{l}{\begin{tabular}[l]{@{}l@{}}Two speakers only;\\hardly applied for \\background noise\end{tabular}} \\\cline{2-4}
			
			& Ephrat et al. \cite{ephrat2018looking} & \begin{tabular}[l]{@{}l@{}}Predict a complex spectrogram \\mask for each speaker; \\trained once, applicable to \\any speaker\end{tabular} & \begin{tabular}[l]{@{}l@{}}The model is too complicated\\and lacks explanation \end{tabular} \\\cline{2-4}
			& Morrone et al. \cite{morrone2019face} & \begin{tabular}[l]{@{}l@{}}  Use landmarks to generate\\time-frequency masks \end{tabular} & \begin{tabular}[l]{@{}l@{}}Additional landmark\\detection required \end{tabular}\\\cline{2-4}
			\hline
			& Gao et al. \cite{gao2018object-sounds} & \begin{tabular}[l]{@{}l@{}}Disentangle audio frequencies \\ related to visual objects\end{tabular} & Separated audio only \\ 
			\cline{2-4}
			& Senocak et al \cite{senocak2018learning} & \begin{tabular}[l]{@{}l@{}}Focus on the primary \\area by using attention \end{tabular} & \begin{tabular}[l]{@{}l@{}}Localized sound \\source only\end{tabular} \\ 
			\cline{2-4}
			& Tian et al. \cite{DBLP:conf/eccv/TianSLDX18} & \begin{tabular}[l]{@{}l@{}} Joint modeling of auditory \\and visual modalities \end{tabular} &  \begin{tabular}[l]{@{}l@{}} Localized sound \\source only \end{tabular} \\\cline{2-4}
			
			\begin{tabular}[l]{@{}l@{}} Separate and Localize \\Objects' Sounds \end{tabular} & Pu et al. \cite{pu2017audio} & \begin{tabular}[l]{@{}l@{}}Use low rank to extract the \\sparsely correlated components \end{tabular} & \begin{tabular}[l]{@{}l@{}}Not for the in-the-wild \\ environment \end{tabular}\\ 
			\cline{2-4}
			& Zhao et al. \cite{zhao2018sound} & \begin{tabular}[l]{@{}l@{}}Mix and separate a given audio;\\without traditional~supervision\end{tabular} & \begin{tabular}[l]{@{}l@{}} Motion information \\is not considered \end{tabular}\\\cline{2-4}
			& Zhao et al. \cite{DBLP:journals/corr/abs-1904-05979} & \begin{tabular}[l]{@{}l@{}} Introduce motion trajectory \\and curriculum learning \end{tabular} &  \begin{tabular}[l]{@{}l@{}}Only suitable for synchronized \\video and audio input \end{tabular} \\\cline{2-4}
			& Rouditchenko et al. \cite{DBLP:journals/corr/abs-1904-09013} & \begin{tabular}[l]{@{}l@{}} Separation and localization use \\only one modality input \end{tabular} &  \begin{tabular}[l]{@{}l@{}} Does not fully utilize\\ temporal information \end{tabular} \\\cline{2-4}
			& Parekh et al. \cite{DBLP:journals/corr/abs-1811-04000} & \begin{tabular}[l]{@{}l@{}} Weakly supervised learning \\via multiple-instance learning  \end{tabular} &  \begin{tabular}[l]{@{}l@{}}Only a bounding box\\proposed on the image \end{tabular} \\\cline{2-4}
			\hline
		\end{tabular}
	\end{table*}
	
	\subsubsection{Localization}
    Instead of only separating audio, can machines localize the sound source merely by observing sound and visual scene pairs as a human can? There is evidence both in physiology and psychology that sound localization of acoustic signals is strongly influenced by synchronicity of their visual signals \cite{Hershey00audio-vision}. 
    The past efforts in this domain were limited to requiring specific devices or additional features. 
    Izadinia et al. \cite{izadinia2013multimodal} proposed utilizing the velocity and acceleration of moving objects as visual features to assign sounds to them. Zunino et al. \cite{zunino2015seeing} presented a new hybrid device for sound and optical imaging that was primarily suitable for automatic monitoring.
	
    As the number of unlabeled videos on the Internet has been increasing dramatically, recent methods mainly focus on unsupervised learning.
    Additionally, modeling audio and visual modalities simultaneously tends to outperform independent modeling. Senocak et al. \cite{senocak2018learning} learned to localize sound sources by merely watching and listening to videos. The relevant model mainly consisted of three networks, namely, sound and visual networks and an attention network trained via the distance ratio \cite{hoffer2015deep} unsupervised loss.
    
    Attention mechanisms cause the model to focus on the primary area. They provide prior knowledge in a semisupervised setting. As a result, the network can be converted into a unified one that can learn better from data without additional annotations.
    To enable cross-modality localization, Tian et al. \cite{DBLP:conf/eccv/TianSLDX18} proposed capturing the semantics of sound-emitting objects via the learned attention and leveraging temporal alignment to discover the correlations between the two modalities.
	
	\subsubsection{ Simultaneous Separation and Localization}
    Sound source separation and localization can be strongly associated with each other by assigning one modality's information to another. Therefore, several researchers attempted to perform localization and separation simultaneously. Pu et al. \cite{pu2017audio} used a low-rank and sparse framework to model the background. The researchers extracted 
components with sparse correlations between the audio and visual modalities. However, the scenario of this method had a major limitation: it could only be applied to videos with a few sound-generating objects.
    Therefore, Zhao et al. \cite{zhao2018sound} introduced a system called PixelPlayer that used a two-stream network and presented a mix-and-separate framework to train the entire network. In this framework, audio signals from two different videos were added to produce a mixed signal as input. The input was then fed into the network that was trained to separate the audio source signals based on the corresponding video frames. The two separated sound signals were treated as outputs. The system thus learned to separate individual sources without traditional supervision.
    
    Instead of merely relying on image semantics while ignoring the temporal motion information in the video, Zhao et al. \cite{DBLP:journals/corr/abs-1904-05979} subsequently proposed an end-to-end network called deep dense trajectory to learn the motion information for audio-visual sound separation. 
    Furthermore, due to the lack of training samples, directly separating sound for a single class of instruments tend to lead to overfitting. Therefore, the authors proposed a curriculum strategy, starting by separating sounds from different instruments and proceeding to sounds from the same instrument. This gradual approach provided a good start for the network to converge better on the separation and localization tasks. 	
	
    
    The methods of previous studies \cite{pu2017audio, zhao2018sound, DBLP:journals/corr/abs-1904-05979} could only be applied to videos with synchronized audio.
    Hence, Rouditchenko et al. \cite{DBLP:journals/corr/abs-1904-09013} tried to perform localization and separation tasks using only video frames or sound by disentangling concepts learned by neural networks. The researchers proposed an approach to produce sparse activations that could correspond to semantic categories in the input using the sigmoid activation function during the training stage and softmax activation during the fine-tuning stage. 
    Afterwards, the researchers assigned these semantic categories to intermediate network feature channels using labels available in the training dataset. In other words, given a video frame or a sound, the approach used the category-to-feature-channel correspondence to select a specific type of source or object for separation or localization.
    Aiming to introduce weak labels to improve performance, Parekh et al. \cite{DBLP:journals/corr/abs-1811-04000} designed an approach based on multiple-instance learning, a well-known strategy for weakly supervised learning.

	\section{Audio-visual Correspondence Learning}	
	\label{sec:corr}
	
    In this section, we introduce several studies that explored the global semantic relation between audio and visual modalities. We name this branch of research ``audio-visual correspondence learning''; it consists of 1) the audio-visual matching task and 2) the audio-visual speech recognition task.

    \subsection{Audio-visual Matching}
    
    Biometric authentication, ranging from facial recognition to fingerprint and iris authentication, is a popular topic that has been researched over many years, while evidence shows that this system can be attacked maliciously. To detect such attacks, recent studies particularly focus on speech antispoofing measures.  
    
    Sriskandaraja et al. \cite{sriskandaraja2018deep} proposed a network based on a Siamese architecture to evaluate the similarities between pairs of speech samples. 
    \cite{bialobrzeski2019robust} presented a two-stream network, where the first network was a Bayesian neural network assumed to be overfitting, and the second network was a CNN used to improve generalization. 
    Alanis et al. \cite{gomez2019light} further incorporated LightCNN \cite{wu2018light} and a gated recurrent unit (GRU) \cite{chung2014empirical} as a robust feature extractor to represent speech signals in utterance-level analysis to improve performance. 
    
    We note that cross-modality matching is a special form of such authentication that has recently been extensively studied. It attempts to learn the similarity between pairs. We divide this matching task into fine-grained voice-face matching and coarse-grained audio-image retrieval. 
	
	\subsubsection{Voice-Facial Matching}
\begin{figure}
		\centering
		\includegraphics[width=1\linewidth]{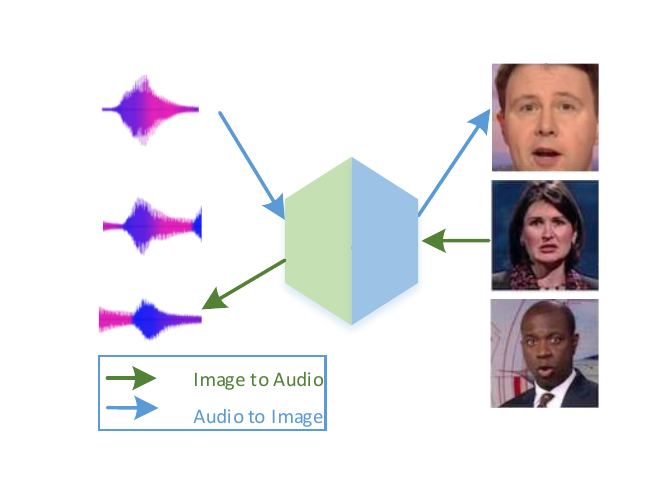}
		\caption{Demonstration of Audio-to-Image retrieval (a) and Image-to-Audio retrieval (b).}
		\label{audio-visual_correspondence_learning}
	\end{figure}

    Given facial images of different identities and the corresponding audio sequences, voice-facial matching aims to identify the face that the audio belongs to (the V2F task) or vice versa (the F2V task), as shown in Fig. \ref{audio-visual_correspondence_learning}. The key point is finding the embedding between audio and visual modalities. 
    Nagrani et al. \cite{DBLP:journals/corr/abs-1804-00326} proposed using three networks to address the audio-visual matching problem: a static network, a dynamic network, and an N-way network. The static network and the dynamic network could only handle the problem with a specific number of images and audio tracks. The difference was that the dynamic network added to each image temporal information such as the optical flow or a 3D convolution \cite{DBLP:journals/corr/TorfiIND17,NIPS2014_5353}. Based on the static network, the authors increased the number of samples to form an N-way network that was able to solve the $N : 1$ identification problem.
    
    However, the correlation between the two modalities was not fully utilized in the above method. Therefore, Wen et al. \cite{2018arXiv180704836W} proposed a disjoint mapping network (DIMNets) to fully use the covariates (e.g., gender and nationality) ~\cite{DBLP:journals/corr/IoffeS15,Lippert10166} to bridge the relation between voice and face information. The intuitive assumption was that
    for a given voice and face pair, the more covariates were shared between the two modalities, the higher the probability of being a match. The main drawback of this framework was that a large number of covariates led to high data costs. 
    Therefore, Hoover et al. \cite{DBLP:journals/corr/HooverCPSS17} suggested a low-cost but robust approach of detection and clustering on audio clips and facial images. 
    For the audio stream, the researchers applied a neural network model to detect speech for clustering and subsequently assigned a frame cluster to the given audio cluster according to the majority principle. Doing so required a small amount of data for pretraining. 
	
    To further enhance the robustness of the network, Chung et al. \cite{2018arXiv180908001C} proposed an improved two-stream training method that increased the number of negative samples to improve the error-tolerance rate of the network. 
    The cross-modality matching task, which is essentially a classification task, allows for wide-ranging applications of the triplet loss. However, it is 
fragile in the case of multiple samples. 
    To overcome this defect, Wang et al. \cite{wang2019novel} proposed a novel loss function to expand the triplet loss for multiple samples and a new elastic network (called Emnet) based on a two-stream architecture that can tolerate a variable number of inputs to increase the flexibility of the network.
	
	\subsubsection{Audio-image Retrieval}
    The cross-modality retrieval task aims to discover the relationship between different modalities. Given one sample in the source modality, the proposed model can retrieve the corresponding sample with the same identity in the target modality. 
    For audio-image retrieval as an example, the aim is to return a relevant piano sound, given a picture of a girl playing a piano. Compared with the previously considered voice and face matching, this task is more coarse-grained.
    
    Unlike other retrieval tasks such as the text-image task \cite{529712,10.1117/12.234775,Rasiwasia:2010:NAC:1873951.1873987} or the sound-text task \cite{DBLP:journals/corr/AytarVT17}, the audio-visual retrieval task mainly focuses on subspace learning.
    Didac et al.~\cite{DBLP:journals/corr/abs-1801-02200} proposed a new joint embedding model that mapped two modalities into a joint embedding space, and then directly calculated the Euclidean distance between them. 
    The authors leveraged cosine similarity to ensure that the two modalities in the same space were as close as possible while not overlapping. 
    Note that the designed architecture would have a large number of parameters due to the existence of a large number of fully connected layers.
    
	Hong et al.~\cite{Hong2017DeepLF} 
	proposed a joint embedding model that relied on pretrained networks and used CNNs to replace fully connected layers to reduce the number of parameters to some extent. 
	The video and music were fed to the pretrained network and then aggregated, followed by a two-stream network trained via the intermodal ranking loss. In addition, to preserve modality-specific characteristics, the researchers proposed a novel soft intramodal structure loss. 
	However, the resulting network was very complex and difficult to apply in practice. To solve this problem, Arsha et al. \cite{DBLP:journals/corr/abs-1805-00833} proposed a cross-modality self-supervised method to learn the embedding of audio and visual information from a video and significantly reduced the complexity of the network. For sample selection, the authors designed a novel curriculum learning schedule to further improve performance. In addition, the resulting joint embedding could be efficiently and effectively applied in practical applications.
	
	
	\begin{table*}[h!]
		\centering
		\label{tab:cross-modal_task_summary}
		\caption{Summary of audio-visual correspondence learning.}	\begin{tabular}{ccll} 
			\hline
			\textbf{Category}  & Method & Ideas \& Strengths & Weaknesses \\ 
			\hline
			& Nagrani et al. ~\cite{DBLP:journals/corr/abs-1804-00326} & \begin{tabular}[l]{@{}l@{}}The method is novel and \\ incorporates dynamic information \end{tabular} & \begin{tabular}[l]{@{}l@{}} As the sample size increases, \\the accuracy decreases excessively \end{tabular} \\ 
			\cline{2-4}
			& Wen et al. ~\cite{2018arXiv180704836W}. & \begin{tabular}[l]{@{}l@{}}The correlation between \\modes is utilized\end{tabular} & \begin{tabular}[l]{@{}l@{}}Dataset acquisition is difficult \end{tabular} \\ 
			\cline{2-4}
			
			Voice-Face Matching & Wang et al. \cite{8794873} & \begin{tabular}[l]{@{}l@{}}Can deal with multiple samples \\ Can change the size of input  \end{tabular} & \multicolumn{1}{l}{\begin{tabular}[l]{@{}l@{}}Static image only;\\model complexity\end{tabular}} \\\cline{2-4}
			& Hoover et al. \cite{DBLP:journals/corr/HooverCPSS17} & \begin{tabular}[l]{@{}l@{}}Easy to implement\\Robust \\ Efficient~\\\end{tabular} & Cannot handle large-scale data \\ 
			\hline
			& Hong et al. \cite{Hong2017DeepLF} & \begin{tabular}[l]{@{}l@{}}Preserve modality- \\specific characteristics \\ Soft intra-modality structure loss \end{tabular} & Complex network \\ 
			\cline{2-4}
			Audio-visual retrieval & Didac et al. \cite{DBLP:journals/corr/abs-1801-02200} & \begin{tabular}[l]{@{}l@{}}Metric Learning \\ Using fewer parameters \end{tabular} & \begin{tabular}[l]{@{}l@{}}Only two faces \\Static images\end{tabular} \\ 
			\cline{2-4}
			
			\cline{2-4}
			& Arsha et al. \cite{DBLP:journals/corr/abs-1805-00833} & \begin{tabular}[l]{@{}l@{}}Curriculum learning\\Applied value\\Low data cost\\\end{tabular} &  Low accuracy for multiple samples\\
			\hline
						& Petridis et al. ~\cite{petridis2017end} & \begin{tabular}[l]{@{}l@{}}Simultaneously obtain \\feature and classification  \end{tabular} & \begin{tabular}[l]{@{}l@{}} Lack of audio information \end{tabular} \\ 
			\cline{2-4}
			& Wand et al. ~\cite{wand2016lipreading}. & \begin{tabular}[l]{@{}l@{}}LSTM\\
				Simple method\end{tabular} & \begin{tabular}[l]{@{}l@{}}Word-level \end{tabular} \\ 		
			\cline{2-4}
			
			Audio-visual Speech Recognition & Shillingford et al. \cite{assael2016lipnet} & \begin{tabular}[l]{@{}l@{}}Sentence-level\\LipNet \\ CTC loss \end{tabular} & \multicolumn{1}{l}{\begin{tabular}[l]{@{}l@{}}No audio information\end{tabular}} \\\cline{2-4}
			& Chung et al. \cite{chung2017lip} & \begin{tabular}[l]{@{}l@{}}Audio and visual information\\LRS dataset~\end{tabular} & Noise is not considered \\ 
			\cline{2-4}
			& Trigeorgis et al. \cite{trigeorgis2016adieu} & \begin{tabular}[l]{@{}l@{}}Audio information\\The algorithm is robust~\end{tabular} & Noise is not considered \\ 
			\cline{2-4}
			& Afouras et al.~\cite{afouras2018deep} & 
			\begin{tabular}[l]{@{}l@{}}Study noise in audio \\ LRS2-BBC Dataset \end{tabular} & Complex network \\ 
			\hline
		\end{tabular}
	\end{table*}

	\begin{figure}
		\centering
		\includegraphics[width=1\linewidth]{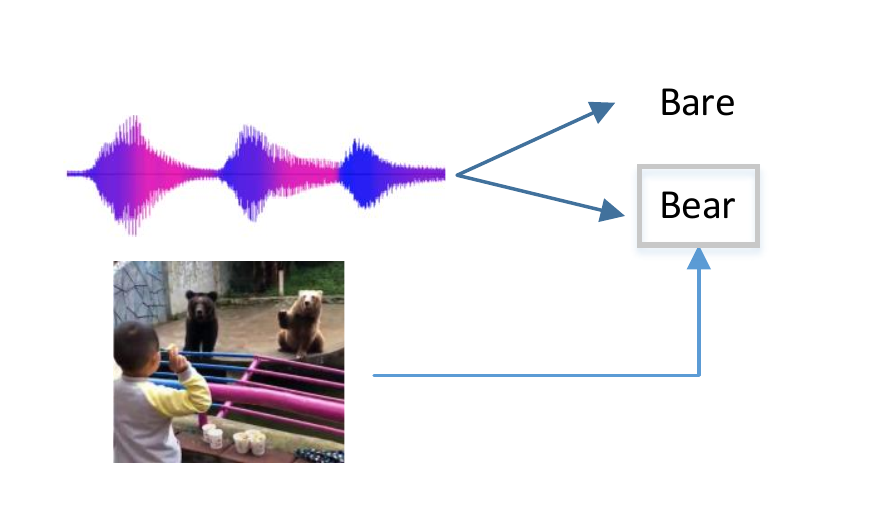}
		\caption{Demonstration of audio-visual speech recognition. }
		\label{fig:audio-visual_recognition}
	\end{figure}
	
	\subsection{Audio-visual Speech Recognition}	
    The recognition of content of a given speech clip has been studied for many years, yet despite great achievements, researchers are still aiming for satisfactory performance in challenging scenarios. 
    Due to the correlation between audio and vision, combining these two modalities tends to offer more prior information. For example, one can predict the scene where the conversation took place, which provides a strong prior for speech recognition, as shown in Fig. \ref{fig:audio-visual_recognition}. 
    
    Earlier efforts on audio-visual fusion models usually consisted of two steps: 1) extracting features from the image and audio signals and 2) combining the features for joint classification ~\cite{dupont2000audio,petridis2016prediction,potamianos2003recent}. Later, taking advantage of deep learning, feature extraction was replaced with a neural network encoder \cite{hu2016temporal, ngiam2011multimodal, ninomiya2015integration}.
    Several recently studies have shown a tendency to use an end-to-end approach to visual speech recognition. These studies can be mainly divided into two groups. 
    They either leverage the fully connected layers and LSTM to extract features and model the temporal information \cite{petridis2017end, wand2016lipreading} or use a 3D convolutional layer followed by a combination of CNNs and LSTMs \cite{assael2016lipnet, stafylakis2017combining}.
    Instead of adopting a two-step strategy, Petridis et al. \cite{petridis2017end} introduced an audio-visual fusion model that simultaneously extracted features directly from pixels and spectrograms and performed classification of speech and nonlinguistic vocalizations. Furthermore, temporal information was extracted by a bidirectional LSTM.
    Although this method could perform feature extraction and classification at the same time, it still followed the two-step strategy. 
    
    To this end, Wand et al. \cite{wand2016lipreading} presented a word-level lip-reading system using LSTM.
    In contrast to previous methods, Assael et.al \cite{assael2016lipnet} proposed a novel end-to-end LipNet model based on sentence-level sequence prediction, which consisted of spatial-temporal convolutions, a recurrent network and a model trained via the connectionist temporal classification (CTC) loss. Experiments showed that lip-reading outperformed the two-step strategy.
    
    However, the limited information in the visual modality may lead to a performance bottleneck. To combine both audio and visual information for various scenes, especially in noisy conditions, Trigeorgis et al. ~\cite{trigeorgis2016adieu} introduced an end-to-end model to obtain a `context-aware' feature from the raw temporal representation.
    
    Chung et al. \cite{chung2017lip} presented a ``Watch, Listen, Attend, and Spell'' (WLAS) network to explain the influence of audio on the recognition task. The model took advantage of the dual attention mechanism and could operate on a single or combined modality. To speed up the training and avoid overfitting, the researchers also used a curriculum learning strategy. 
    To analyze an ``in-the-wild'' dataset, Cui et al.~\cite{sssss} proposed another model based on residual networks and a bidirectional GRU \cite{chung2014empirical}.
    However, the authors did not take the ubiquitous noise in the audio into account.
    To solve this problem, Afouras et al.~\cite{afouras2018deep} proposed a model for performing speech recognition tasks. The researchers compared two common sequence prediction types: connectionist temporal classification and sequence-to-sequence (seq2seq) methods in their models. 
    In the experiment, they observed that the model using seq2seq could perform better according to word error rate (WER) when it was only provided with silent videos. 
    For pure-audio or audio-visual tasks, the two methods behaved similarly. In a noisy environment, the performance of the seq2seq model was worse than that of the corresponding CTC model, suggesting that the CTC model could better handle background noises. 

    \section{Audio and Visual Generation}
    \label{sec:generation} 
    The previously introduced retrieval task shows that the trained model is able to find the most similar audio or visual counterpart.
    While humans can imagine the scenes corresponding to sounds, and vice versa, researchers have tried to endow machines with this kind of imagination for many years.
    Following the invention and advances of generative adversarial networks (GANs) \cite{goodfellow2014generative}, image or video generation has emerged as a topic. It involves several subtasks, including generating images or video from a potential space \cite{arjovsky2017wasserstein}, cross-modality generation \cite{chen2017deep,zhu2018high}, etc. These applications are also relevant to other tasks, e.g., domain adaptation \cite{wei2018person,huang2018auggan}.
    Due to the difference between audio and visual modalities, the potential correlation between them is nonetheless difficult for machines to discover. Generating sound from a visual signal or vice versa, therefore, becomes a challenging task.
    
\begin{figure}
		\centering
		\subfigure[Demonstration of generating speech from lip sequences]{
			\includegraphics[width=1\linewidth]{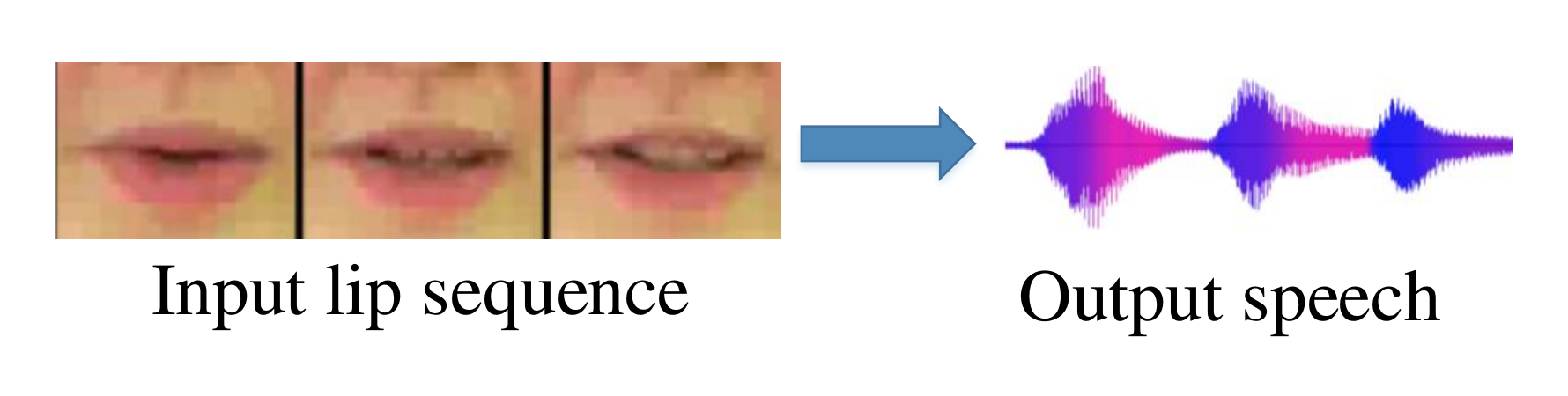}
		}
		\subfigure[Demonstration of video-to-audio generation]{
			\includegraphics[width=1\linewidth]{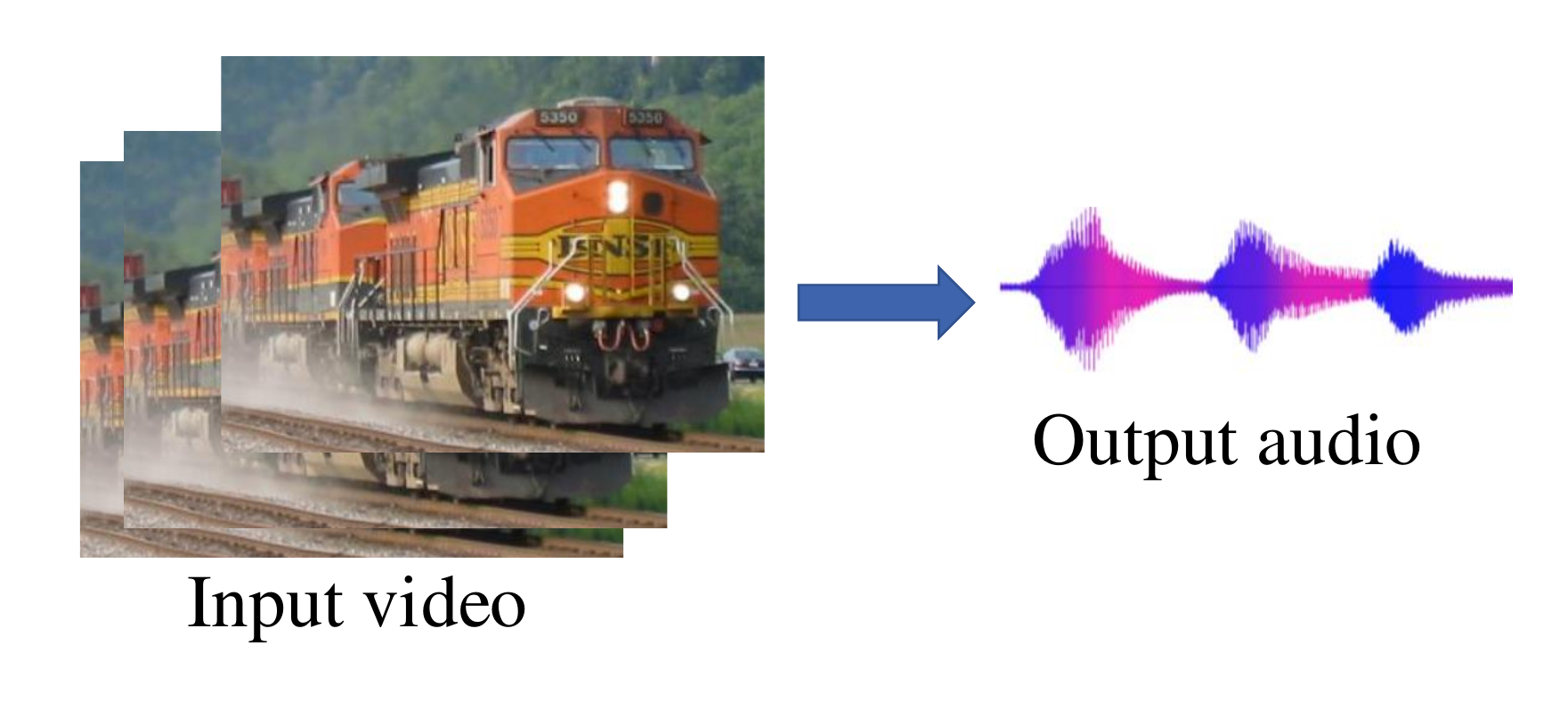}
		}
		
		\caption{Demonstration of visual-to-audio generation.}
		\label{audio-visual_generation}
	\end{figure}

\begin{figure}[hbt]
		\centering
		\subfigure[Demonstration of audio-to-images generation.]{
			\includegraphics[width=1\linewidth]{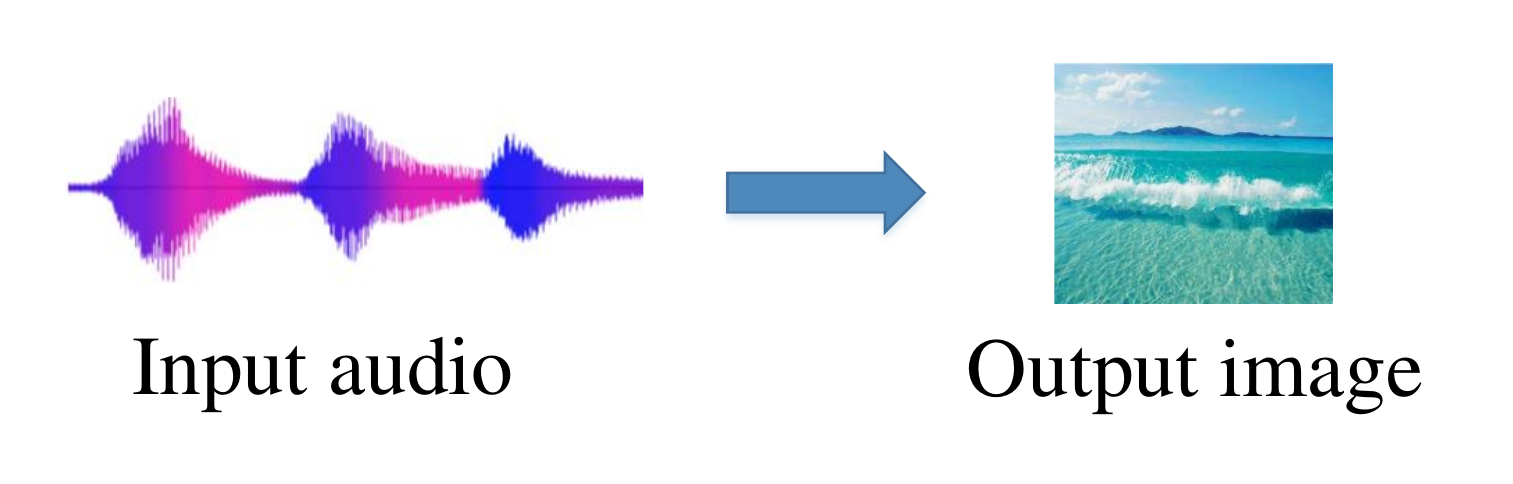}
		}
		\subfigure[Demonstration of a moving body.]{
			\includegraphics[width=1\linewidth]{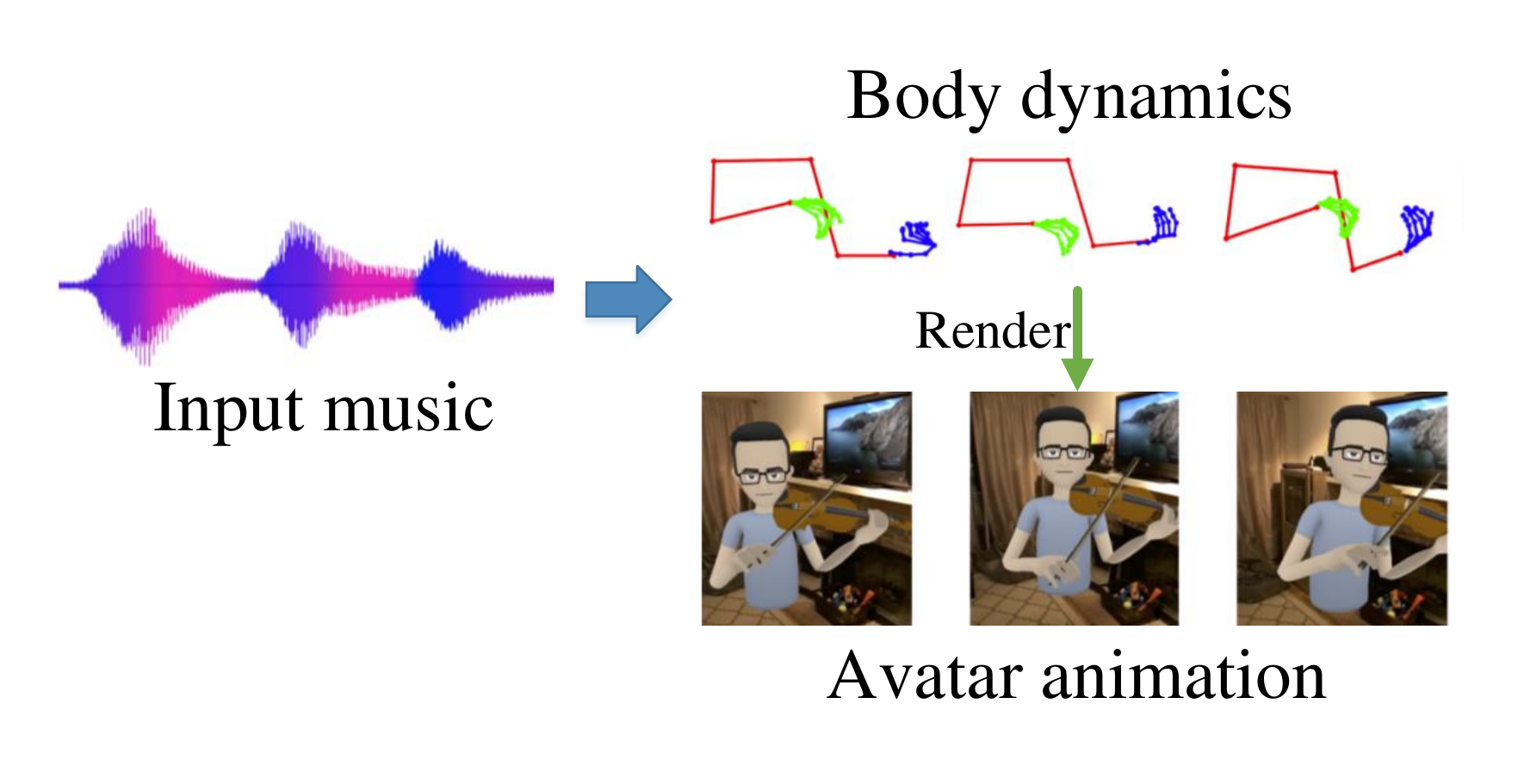}
		}
		\subfigure[Demonstration of a talking face.]{
			\includegraphics[width=1\linewidth]{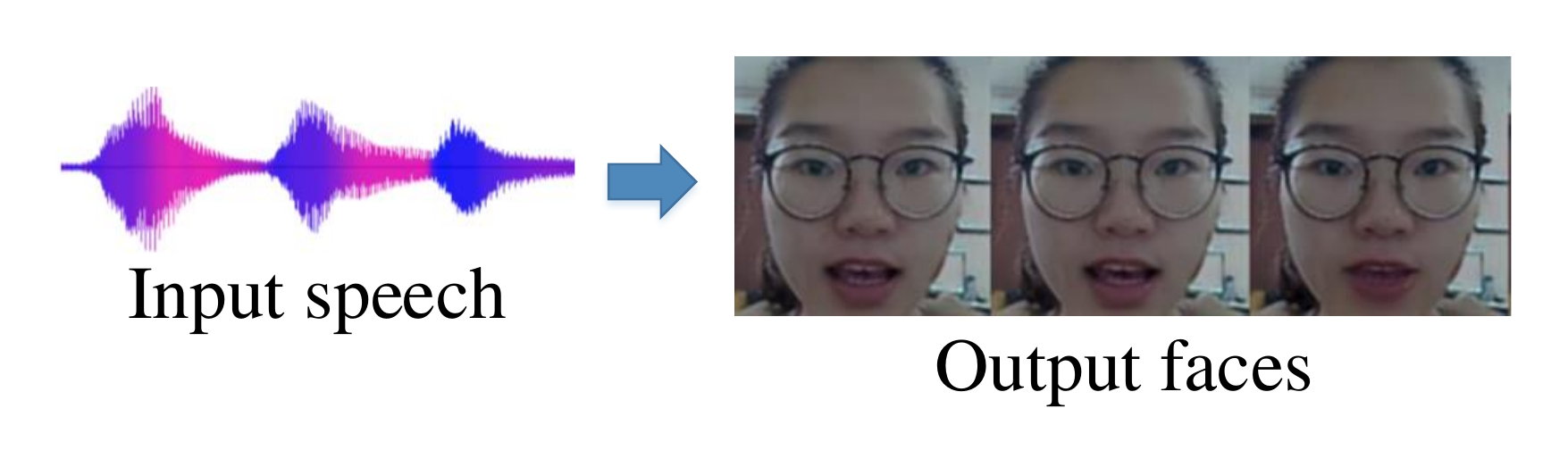}
		}
		\caption{Demonstration of talking face generation and moving body generation.}
		\label{fig:audio2video}
	\end{figure}

    In this section, we will mainly review the recent development of audio and visual generation, i.e., generating audio from visual signals or vice versa. Visual signals here mainly refer to images, motion dynamics, and videos. 
    The subsection `Visual to Audio' mainly focuses on recovering the speech from the video of the lip area (Fig. \ref{audio-visual_generation} (a)) or generating sounds that may occur in the given scenes (Fig. \ref{audio-visual_generation} (a)). 
    In contrast, the discussion of `Audio to Visual' generation (Fig. \ref{audio-visual_generation} (b)) will examine generating images from a given audio (Fig. \ref{fig:audio2video} (a)), body motion generation (Fig. \ref{fig:audio2video} (b)), and talking face generation (Fig. \ref{fig:audio2video} (c)).
		\begin{table*}[h]
		\centering
		\caption{Summary of recent approaches to video-to-audio generation.}
		\label{tab:visual2audio_generation_summary}
		\begin{tabular}{ccll} 
			\hline
			\textbf{Category}  & Method & Ideas \& Strengths & Weaknesses \\ 
			\hline
			& Cornu et al. \cite{cornu2015reconstructing} & \begin{tabular}[l]{@{}l@{}} Reconstruct intelligible \\speech only from\\ visual speech features\end{tabular}  & \begin{tabular}[l]{@{}l@{}} Applied to limited scenarios \end{tabular} \\ 
			\cline{2-4}
			\begin{tabular}[l]{@{}l@{}}Lip sequence \\to Speech\end{tabular} & Ephrat et al. \cite{ephrat2017improved}  & \begin{tabular}[l]{@{}l@{}} Compute optical flow\\between frames\end{tabular}  & Applied to limited scenarios \\ 
			\cline{2-4}
			& Cornu et al. \cite{le2017generating}   & \begin{tabular}[l]{@{}l@{}} Reconstruct speech using\\ a classification approach\\combined with feature-level\\temporal information\end{tabular} & \begin{tabular}[l]{@{}l@{}}Cannot apply to real-time\\ conversational speech\end{tabular} \\ 
			\hline
			& Davis et al. \cite{davis2014visual} & \begin{tabular}[l]{@{}l@{}}Recover real-world audio by \\capturing vibrations of objects \end{tabular} & \begin{tabular}[l]{@{}l@{}}Requires a specific device; \\ can only be applied to \\ soft objects \end{tabular} \\ 
			\cline{2-4}
			& Owens et al. \cite{owens2016visually} & \begin{tabular}[l]{@{}l@{}}Use LSTM to capture \\ the relation between material \\ and motion \end{tabular} & \begin{tabular}[l]{@{}l@{}} For a lab-controlled  \\ environment only \end{tabular} \\ 
			\cline{2-4}
			\begin{tabular}[l]{@{}l@{}}General Video\\to Audio \end{tabular}& Zhou et al. \cite{zhou2017visual} & \begin{tabular}[l]{@{}l@{}}Leverage a hierarchical \\RNN~to generate \\ in-the-wild~sounds~\end{tabular} & Monophonic audio only \\ 
			\cline{2-4}
			& Morgado et al. \cite{morgado2018self} & \begin{tabular}[l]{@{}l@{}}Localize and \\ separate sounds to \\generate spatial audio~\\from 360$^\circ$ video\end{tabular} & \begin{tabular}[l]{@{}l@{}}Fails sometimes;\\ 360$^\circ$ video required \end{tabular} \\ 
			\hline
			
		\end{tabular}
	\end{table*}

    \subsection{Vision-to-Audio Generation}
    Many methods have been explored to extract audio information from visual information, including predicting sounds from visually observed vibrations and generating audio via a video signal. We divide the visual-to-audio generation tasks into two categories: generating speech from lip video and synthesizing sounds from general videos without scene limitations.
    
    \subsubsection{Lip Sequence to Speech}
    There is a natural relationship between speech and lips. Separately from understanding the content of speech by observing lips (lip-reading), several studies have tried to reconstruct speech by observing lips. 
    Cornu et al. \cite{cornu2015reconstructing} attempted to predict the spectral envelope from visual features, combining it with artificial excitation signals, and to synthesize audio signals in a speech production model.
    Ephrat et al. \cite{ephrat2017vid2speech} proposed an end-to-end model based on a CNN to generate audio features for each silent video frame based on its adjacent frames. The waveform was therefore reconstructed based on the learned features to produce understandable speech.

    Using temporal information to improve speech reconstruction has been extensively explored. Ephrat et al. \cite{ephrat2017improved} proposed leveraging the optical flow to capture the temporal motion at the same time. 
    Cornu et al. \cite{le2017generating} leveraged recurrent neural networks to incorporate temporal information into the prediction.
    
    \subsubsection{General Video to Audio}
    When a sound hits the surfaces of some small objects, the latter will vibrate slightly. Therefore, Davis et al. \cite{davis2014visual} utilized this specific feature to recover the sound from vibrations observed passively by a high-speed camera. Note that it should be easily for suitable objects to vibrate, which is the case for a glass of water, a pot of plants, or a box of napkins.
    We argue that this work is similar to the previously introduced speech reconstruction studies \cite{cornu2015reconstructing, ephrat2017vid2speech, ephrat2017improved, le2017generating} since all of them use the relation between visual and sound context. In speech reconstruction, the visual part concentrates more on lip movement, while in this work, it focuses on small vibrations.
    
    Owens et al. \cite{owens2016visually} observed that when different materials were hit or scratched, they emitted a variety of sounds. Thus, the researchers introduced a model that learned to synthesize sound from a video in which objects made of different materials were hit with a drumstick at different angles and velocities. The researchers demonstrated that their model could not only identify different sounds originating from different materials but also learn the pattern of interaction with objects (different actions applied to objects result in different sounds). The model leveraged an RNN to extract sound features from video frames and subsequently generated waveforms through an instance-based synthesis process.
    
    Although Owens et al. \cite{owens2016visually} could generate sound from various materials, the authors' approach still could not be applied to real-life applications since the network was trained by videos shot in a lab environment under strict constraints. To improve the result and generate sounds from in-the-wild videos, Zhou et al. \cite{zhou2017visual} designed an end-to-end model. It was structured as a video encoder and a sound generator to learn the mapping from video frames to sounds. Afterwards, the network leveraged a hierarchical RNN \cite{mehri2016samplernn} for sound generation. Specifically, the authors trained a model to directly predict raw audio signals (waveform samples) from input videos. They demonstrated that this model could learn the correlation between sound and visual input for various scenes and object interactions.

    The previous efforts we have mentioned focused on monophonic audio generation, while Morgado et al. \cite{morgado2018self} attempted to convert monophonic audio recorded by a 360$^{\circ}$ video camera into spatial audio. Performing such a task of audio specialization requires addressing two primary issues: source separation and localization.
    Therefore, the researchers designed a model to separate the sound sources from mixed-input audio and then localize them in the video. Another multimodality model was used to guide the separation and localization since the audio and video were complementary.

    \subsection{Audio to Vision}
    In this section, we provide a detailed review of audio-to-visual generation. 
    We first introduce audio-to-images generation, which is easier than video generation since it does not require temporal consistency between the generated images.
    
    \subsubsection{Audio to Image}
    To generate images of better quality, Wan et al. \cite{wan2018towards} put forward a model that combined the spectral norm, an auxiliary classifier, and a projection discriminator to form the researchers' conditional GAN model. The model could output images of different scales according to the volume of the sound, even for the same sound.
    Instead of generating real-world scenes of the sound that had occurred, Qiu et al. \cite{DBLP:conf/cvpr/QiuK18} suggested imagining the content from music. The authors extracted features by feeding the music and images into two networks and learning the correlation between those features and finally generated images from the learned correlation.
    
    Several studies have focused on audio-visual mutual generation. Chen et al. \cite{chen2017deep} were the first to attempt to solve this cross-modality generation problem using conditional GANs. The researchers defined a sound-to-image (S2I) network and an image-to-sound (I2S) network that generated images and sounds, respectively.
    Instead of separating S2I and I2S generation, Hao et al. \cite{hao2017cmcgan} combined the respective networks into one network by considering a cross-modality cyclic generative adversarial network (CMCGAN) for the cross-modality visual-audio mutual generation task. Following the principle of cyclic consistency, CMCGAN consisted of four subnetworks: audio-to-visual, visual-to-audio, audio-to-audio, and visual-to-visual.
	
    Most recently, some studies have tried to reconstruct facial images from speech clips. Duarte et al. \cite{duarte2019speech} synthesized facial images containing expressions and poses through the GAN model. Moreover, the authors enhanced their model's generation quality by searching for the optimal input audio length.
    To better learn normalized faces from speech, Oh et al. \cite{oh2019speech2face} explored a reconstructive model. The researchers trained an audio encoder by learning to align the feature space of speech with a pretrained face encoder and decoder.
	
\begin{table*}[!h]
	\centering
	\caption{Summary of recent studies of audio-to-visual generation.}
	\label{tab:audio2visual_generation_summary}
	\begin{tabular}{ccll} 
		\hline
		\textbf{Category}  & Method                                                                               & Ideas \& Strengths                                                                                                                & Weaknesses                                                                       \\ 
		\cline{2-4}
		& Wan et al. \cite{wan2018towards}                                                                            & \begin{tabular}[c]{@{}l@{}}Combined many existing techniques\\ to form a GAN \end{tabular}                                          & Low quality                                                                      \\ 
		\cline{2-4}
		& Qiu et al. \cite{DBLP:conf/cvpr/QiuK18}                                                                            & \begin{tabular}[c]{@{}l@{}} Generated images \\ related to music \end{tabular}                                      & Low quality                                                                      \\ 
		\cline{2-4}
		Audio to Image       & Chen et al.       \cite{chen2017deep}                                                                     & \begin{tabular}[c]{@{}l@{}}Generated both audio-to-visual and \\ visual-to-audio models \end{tabular}                                   & \begin{tabular}[c]{@{}l@{}}The models were\\ independent \end{tabular}            \\ 
		\cline{2-4}
		& Hao et al.       \cite{hao2017cmcgan}                                                                     & \begin{tabular}[c]{@{}l@{}} Proposed a cross-modality cyclic \\generative adversarial network \end{tabular}                     & Generated images only                                                              \\ 
		\hline
		& Alemi et al. \cite{alemi2017groovenet}                                                                      & \begin{tabular}[c]{@{}l@{}}Generated dance movements from\\music via real-time GrooveNet \end{tabular}                           &                                                                                  \\ 
		\cline{2-3}
		& Lee et al. \cite{DBLP:journals/corr/abs-1811-00818}                                                                          & \begin{tabular}[c]{@{}l@{}} Generated a choreography system\\via an autoregressive\\encoder-decoder network \end{tabular}           &                                                                                  \\ 
		\cline{2-3}
		Audio to Motions     & Shlizerman et al. \cite{shlizerman2018audio}                                                                    & \begin{tabular}[c]{@{}l@{}} Applied a ``target delay'' LSTM \\to predict body keypoints \end{tabular}                                & \begin{tabular}[c]{@{}l@{}}Constrained to~\\the given dataset\end{tabular}        \\ 
		\cline{2-3}
		& Tang et al. \cite{DBLP:conf/mm/TangJM18} & \begin{tabular}[c]{@{}l@{}}Developed a music-oriented dance \\choreography synthesis method \end{tabular}                                  &                                                                                  \\ 
		\cline{2-3}
		& Yalta et al. \cite{DBLP:journals/corr/abs-1807-01126} & \begin{tabular}[c]{@{}l@{}}Produced weak labels from\\motion directions for\\motion-music alignment \end{tabular}                 &                                                                                  \\ 
		\hline
		& \begin{tabular}[c]{@{}c@{}}Kumar et al. \cite{kumar2017obamanet} and\\ Supasorn et al. \cite{suwajanakorn2017synthesizing} \end{tabular} & \begin{tabular}[c]{@{}l@{}}Generated keypoints \\ by a time-delayed \\ LSTM \end{tabular}                                         & \begin{tabular}[c]{@{}l@{}}Needed retraining for \\ another identity \end{tabular}    \\ 
		\cline{2-4}
		& Chung et al. \cite{Chung2017YouST} & \begin{tabular}[c]{@{}l@{}}Developed an encoder-decoder \\ CNN model suitable\\for more identities \end{tabular}                             &                                                                                  \\ 
		\cline{2-3}
		& Jalalifar et al. \cite{Jalalifar2018SpeechDrivenFR}                                                                 & \begin{tabular}[c]{@{}l@{}}Combined RNN and GAN\\ and applied keypoints \end{tabular}                                            & \begin{tabular}[c]{@{}l@{}}For a lab-controlled \\ environment only \end{tabular}  \\ 
		\cline{2-3}
		Talking Face         & Vougioukas et al.  \cite{Vougioukas2018EndtoEndSF}                                                                   & \begin{tabular}[c]{@{}l@{}}Applied a temporal GAN for \\more temporal
 consistency \end{tabular}                     &                                                                                  \\ 
		\cline{2-4}
		& Chen et al. \cite{Chen2018LipMG}                                                                          & Applied optical flow                                                                                                             & Generated lips only                                                                \\ 
		\cline{2-4}
		& Zhou et al. \cite{Zhou2018TalkingFG}                                                                     & Disentangled information                                                                                                        & Lacked realism                                                                \\ 
		\cline{2-4}
		& Zhu et al. \cite{zhu2018high}                                                                            & \begin{tabular}[c]{@{}l@{}}Asymmetric mutual information estimation \\to capture modality coherence \end{tabular}              & \begin{tabular}[c]{@{}l@{}}Suffered from the ``zoom-in\\ -and-out'' condition \end{tabular}    \\ 
		\cline{2-4}
		& Chen et al. \cite{chen2019hierarchical}                                                                          & Dynamic pixelwise loss                                                                                                        & \begin{tabular}[c]{@{}l@{}}Required multistage \\training \end{tabular}         \\ 
		\cline{2-4}
		& Wiles et al. \cite{Wiles2018X2FaceAN}                                                                          & \begin{tabular}[c]{@{}l@{}}Self-supervised model for\\multimodality driving \end{tabular} & Relatively low quality                                                             \\
		\hline
	\end{tabular}
\end{table*}

	\subsubsection{Body Motion Generation}
	Instead of directly generating videos, numerous studies have tried to animate avatars using motions.
	The motion synthesis methods leveraged multiple techniques, such as dimensionality reduction \cite{DBLP:journals/ijsr/SamadaniKGK13, DBLP:conf/mig/TilmanneD10}, hidden Markov models \cite{DBLP:conf/siggraph/BrandH00}, Gaussian processes \cite{DBLP:conf/icml/WangFH07}, and neural networks \cite{DBLP:conf/icml/TaylorH09, DBLP:journals/corr/Crnkovic-FriisC16, holden2016deep}.
	
	
	
    Alemi et al. \cite{alemi2017groovenet} proposed a real-time GrooveNet based on conditional restricted Boltzmann machines and recurrent neural networks to generate dance movements from music. 
    Lee et al. \cite{DBLP:journals/corr/abs-1811-00818} utilized an autoregressive encoder-decoder network to generate a choreography system from music. 
    Shlizerman et al. \cite{shlizerman2018audio} further introduced a model that used a ``target delay'' LSTM to predict body landmarks. The latter was further used as agents to generate body dynamics. The key idea was to create an animation from the audio that was similar to the action of a pianist or a violinist. In summary, the entire process generated a video of artists' performance corresponding to input audio.
    
    Although previous methods could generate body motion dynamics, the intrinsic beat information of the music has not been used. Tang et al. \cite{DBLP:conf/mm/TangJM18} proposed a music-oriented dance choreography synthesis method that extracted a relation between acoustic and motion features via an LSTM-autoencoder model. Moreover, to achieve better performance, the researchers improved their model with a masking method and temporal indexes.
    Providing weak supervision, Yalta et al. \cite{DBLP:journals/corr/abs-1807-01126} explored producing weak labels from motion direction for motion-music alignment. The authors generated long dance sequences via a conditional autoconfigured deep RNN that was fed by audio spectrum.
	
	
	\subsubsection{Talking Face Generation}
    Exploring audio-to-video generation, many researchers showed great interest in synthesizing people's faces from speech or music. This has many applications, such as animating movies, teleconferencing, talking agents and enhancing speech comprehension while preserving privacy. 
    Earlier studies of talking face generation mainly synthesized a specific identity from the dataset based on an audio of arbitrary speech. Kumar et al. \cite{kumar2017obamanet} attempted to generate key points synced to audio by utilizing a time-delayed LSTM \cite{graves2005framewise} and then generated the video frames conditioned on the key points by another network. Furthermore, Supasorn et al. \cite{suwajanakorn2017synthesizing} proposed a ``teeth proxy'' to improve the visual quality of teeth during generation.
    
    Subsequently, Chung et al. \cite{Chung2017YouST} attempted to use an encoder-decoder CNN model to learn the correspondences between raw audio and videos. Combining RNN and GAN \cite{goodfellow2014generative}, Jalalifar et al. \cite{Jalalifar2018SpeechDrivenFR} produced a sequence of realistic faces that were synchronized with the input audio by two networks. One was an LSTM network used to create lip landmarks out of audio input. The other was a conditional GAN (cGAN) used to generate the resulting faces conditioned on a given set of lip landmarks. Instead of applying cGAN, \cite{Vougioukas2018EndtoEndSF} proposed using a temporal GAN \cite{saito2017temporal} to improve the quality of synthesis. However, the above methods were only applicable to synthesizing talking faces with identities limited to those in a dataset.
    
    Synthesis of talking faces of arbitrary identities has recently drawn significant attention. Chen et al. \cite{Chen2018LipMG} considered correlations among speech and lip movements while generating multiple lip images. The researchers used the optical flow to better express the information between the frames. The fed optical flow represented not only the information of the current shape but also the previous temporal information.
    
    A frontal face photo usually has both identity and speech information. Assuming this, Zhou et al. \cite{Zhou2018TalkingFG} used an adversarial learning method to disentangle different types of information of one image during generation. The disentangled representation had a convenient property that both audio and video could serve as the source of speech information for the generation process. As a result, it was possible to not only output the features but also express 
them more explicitly while applying the resulting network. 
    
    Most recently, to discover the high-level correlation between audio and video, Zhu et al. \cite{zhu2018high} proposed a mutual information approximation to approximate mutual information between modalities. 
    Chen et al. \cite{chen2019hierarchical} applied landmark and motion attention to generating talking faces. The authors further proposed a dynamic pixelwise loss for temporal consistency.
    Facial generation is not limited to specific modalities such as audio or visual since the crucial point is whether there is a mutual pattern between these different modalities. Wiles et al. \cite{Wiles2018X2FaceAN} put forward a self-supervising framework called X2Face to learn the embedded features and generate target facial motions. It could produce videos from any input as long as embedded features were learned.
	
	\section{Audio-visual Representation Learning}
	\label{sec:representation_learning}
	\begin{table*}[!h]
		\centering
		\label{tab:repersentation_learning_summary}
		\caption{Summary of recent audio-visual representation learning studies.}
		\begin{tabular}{llll} 
			\hline
			Type & Method & Ideas \& Strengths & Weaknesses \\ 
			\hline
			\begin{tabular}[l]{@{}l@{}} Single \\modality \end{tabular} & \begin{tabular}[l]{@{}l@{}} Aytar et al. \cite{aytar2016soundnet} \end{tabular} & \begin{tabular}[l]{@{}l@{}}Student-teacher training \\procedure with natural\\video synchronization\end{tabular}  & \begin{tabular}[l]{@{}l@{}} Only learned the \\audio representation \end{tabular} \\ \hline
			& \begin{tabular}[l]{@{}l@{}}Leidal et al.  \cite{Leidal2017LearningMR} \end{tabular} & \begin{tabular}[l]{@{}l@{}}Regularized the amount of \\information encoded in the \\semantic embedding \end{tabular} & \begin{tabular}[l]{@{}l@{}}Focused on spoken utterances\\and handwritten digits \end{tabular} \\ 
			\cline{2-4}		
			& \begin{tabular}[l]{@{}l@{}}Arandjelovic et al. \\ \cite{arandjelovic2017look,arandjelovic2017objects} \end{tabular} & Proposed the AVC task & \begin{tabular}[l]{@{}l@{}} Considered only audio and \\ video correspondence \end{tabular} \\ 
			\cline{2-4}
			\begin{tabular}[l]{@{}l@{}} Dual \\modalities \end{tabular} & Owens et al. \cite{korbar2018co} & \begin{tabular}[l]{@{}l@{}}Proposed the AVTS task\\with curriculum learning\end{tabular} & \begin{tabular}[l]{@{}l@{}}The sound source has to \\feature in the video; only \\one sound source \end{tabular} \\ \cline{2-4}
			& Parekh et al. \cite{parekh2018weakly} & \begin{tabular}[l]{@{}l@{}}Use video labels for weakly\\supervised learning \end{tabular} & \begin{tabular}[l]{@{}l@{}} Leverage the prior knowledge\\of event classification\end{tabular} \\ \cline{2-4}
			& Hu et al. \cite{hu2018deep} & \begin{tabular}[l]{@{}l@{}}Disentangle each \\ modality into a set \\ of distinct components \end{tabular} & \begin{tabular}[l]{@{}l@{}} Require a predefined \\ number of clusters \end{tabular} \\ \hline
		\end{tabular}
	\end{table*}
    Representation learning aims to discover the pattern representation from data automatically. It is motivated by the fact that the choice of data representation usually greatly impacts performance of machine learning \cite{bengio2013representation}. However, real-world data such as images, videos and audio are not amenable to defining specific features algorithmically. 
    
    Additionally, the quality of data representation usually determines the success of machine learning algorithms. Bengio et al.\cite{bengio2013representation} assumed the reason for this to be that different representations could better explain the laws underlying data, and the recent enthusiasm for AI has motivated the design of more powerful representation learning algorithms to achieve these priors.
    
\begin{figure}
		\centering
		\includegraphics[width=1\linewidth]{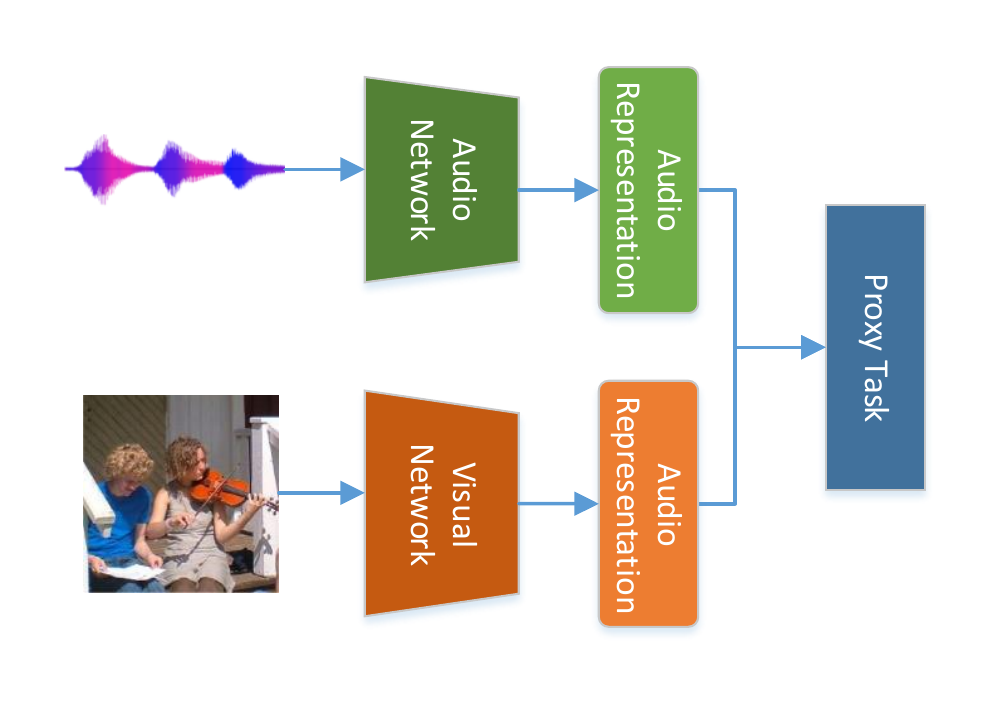}
		\caption{Basic pipeline of representation learning.}
		\label{representation_learning}
	\end{figure}

    In this section, we will review a series of audio-visual learning methods ranging from single-modality \cite{aytar2016soundnet} to dual-modality representation learning \cite{arandjelovic2017objects, arandjelovic2017look, korbar2018co, Leidal2017LearningMR, hu2018deep}. The basic pipeline of such studies is shown in Fig. \ref{representation_learning}.
    
	%
	\subsection{Single-Modality Representation Learning}
	Naturally, to determine whether audio and video are related to each other, researchers focus on determining whether audio and video are from the same video or whether they are synchronized in the same video. Aytar et al. \cite{aytar2016soundnet} exploited the natural synchronization between video and sound to learn an acoustic representation of a video. 
	The researchers proposed a student-teacher training process that used an unlabeled video as a bridge to transfer discernment knowledge from a sophisticated visual identity model to the sound modality. Although the proposed approach managed to learn audio-modality representation in an unsupervised manner, discovering audio and video representations simultaneously remained to be solved.
	
	\subsection{Learning an Audio-visual Representation}
	
    In the corresponding audio and images, the information concerning modality tends to be noisy, while we only require semantic content rather than the exact visual content. 
    Leidal et al. \cite{Leidal2017LearningMR} explored unsupervised learning of the semantic embedded space, which required a close distribution of the related audio and image.  
    The researchers proposed a model to map an input to vectors of the mean and the logarithm of variance of a diagonal Gaussian 
distribution, and the sample semantic embeddings were drawn from these vectors.
    
\begin{figure}[t]
		\centering
		\subfigure[Introduction to the AVC task]{
			\includegraphics[width=1\linewidth]{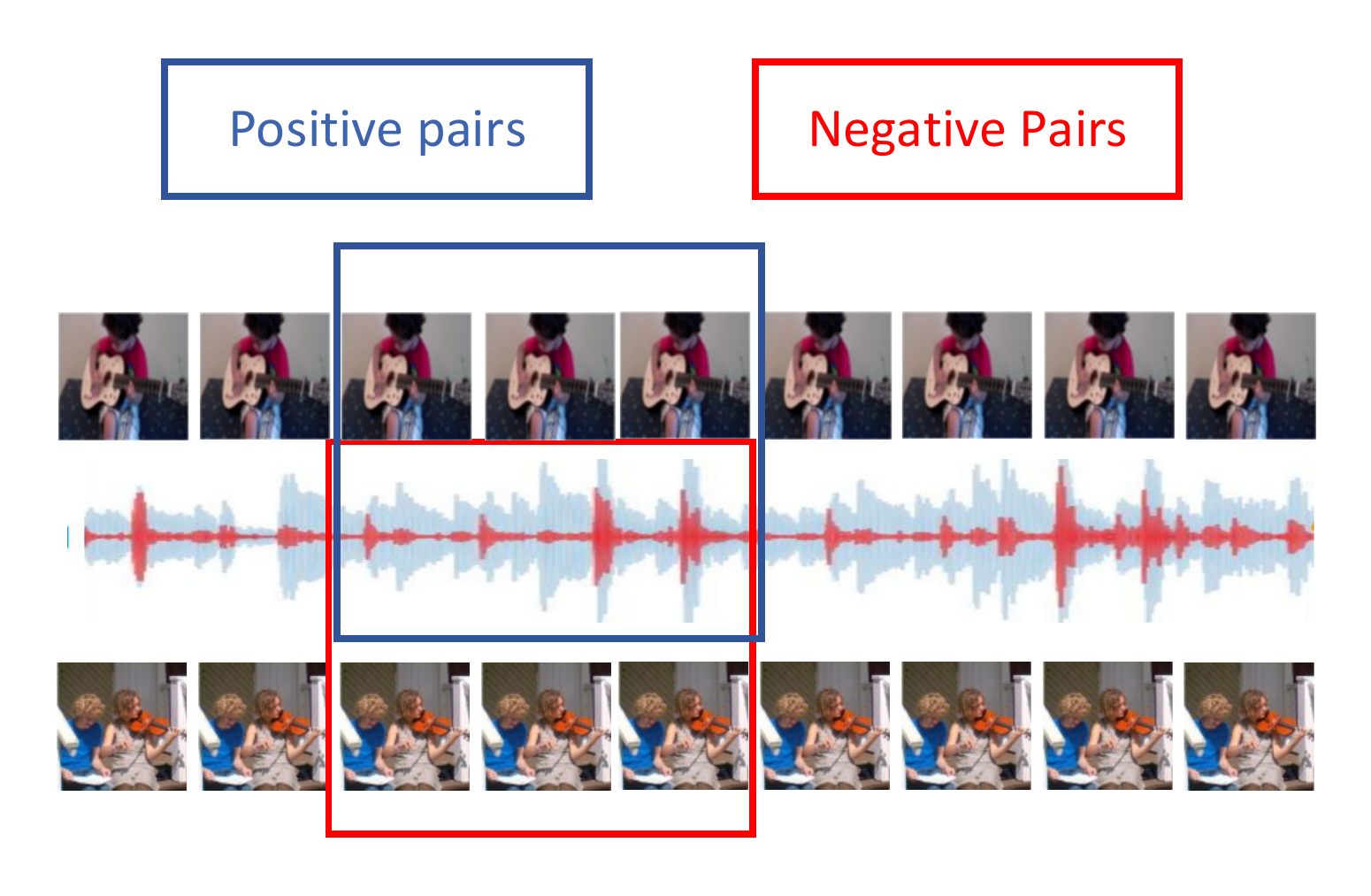}
		}
		\subfigure[Introduction to the AVTS task]{
			\includegraphics[width=1\linewidth]{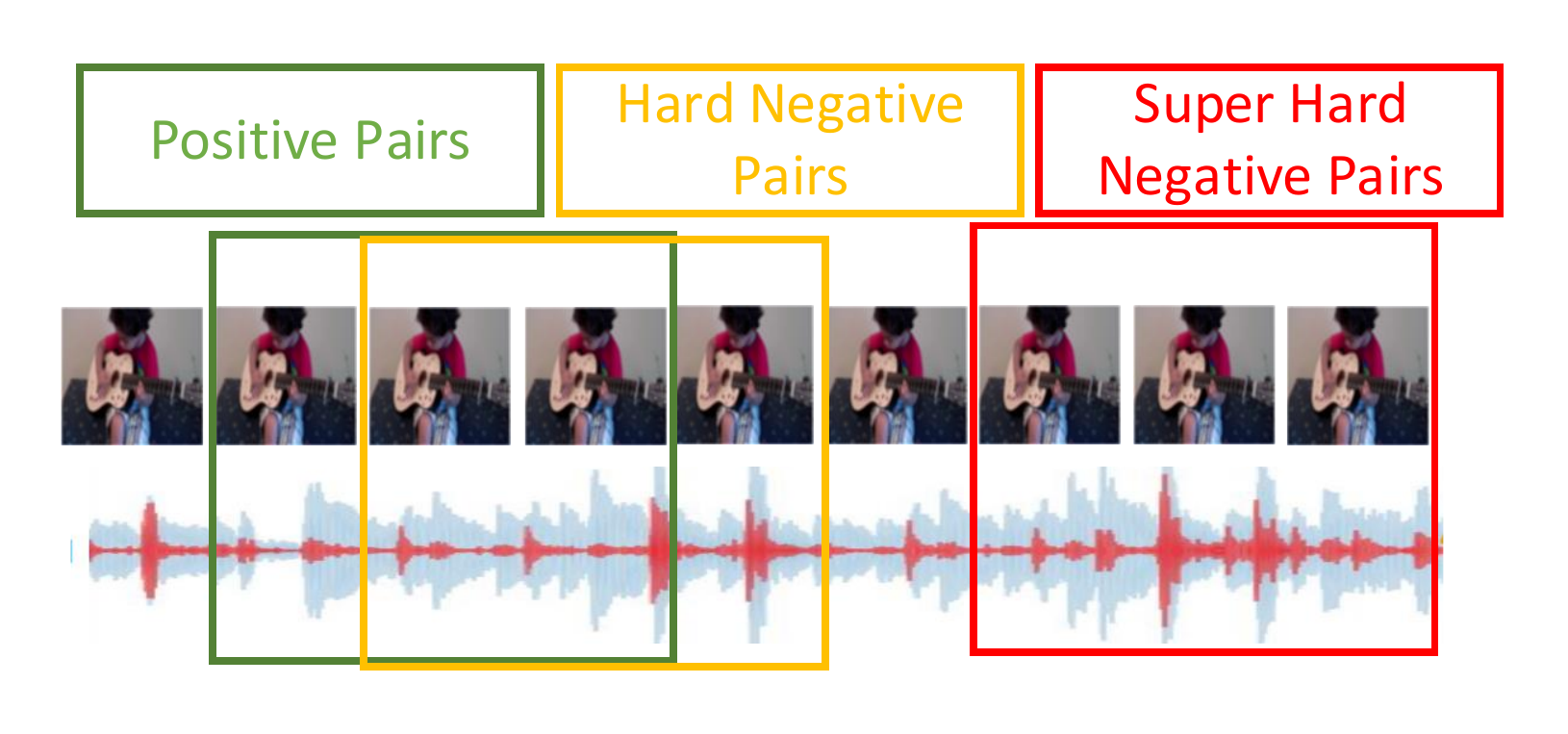}
		}
		\caption{Introduction to the representation task}
		\label{fig:representation-intro}
	\end{figure}

    To learn audio and video's semantic information by simply watching and listening to a large number of unlabeled videos, Arandjelovic et al. \cite{arandjelovic2017look} introduced an audio-visual correspondence learning task (AVC) for training two (visual and audio) networks from scratch, as shown in Fig. \ref{fig:representation-intro} (a).  In this task, the corresponding audio and visual pairs (positive samples) were obtained from the same video, while mismatched (negative) pairs were extracted from different videos. 
    To solve this task, the authors proposed an $L^3$-Net that detected whether the semantics in visual and audio fields were consistent. Although this model was trained without additional supervision, it could learn representations of dual modalities effectively.
	
    Exploring the proposed audio-visual coherence (AVC) task, Arandjelovic et al. \cite{arandjelovic2017objects} continued to investigate AVE-Net that aimed at finding the most similar visual area to the current audio clip. Owens et al. \cite{owens2018audio} proposed adopting a model similar to that of \cite{arandjelovic2017look} but used a 3D convolution network for the videos instead, which could capture the motion information for sound localization.
    
    In contrast to previous AVC task-based solutions, Korbar et al. \cite {korbar2018co} introduced another proxy task called audio-visual time synchronization (AVTS) that further considered whether a given audio sample and video clip were ``synchronized'' or ``not synchronized.''
    In previous AVC tasks, negative samples were obtained as audio and visual samples from different videos. However, exploring AVTS, the researchers trained the model using ``harder'' negative samples representing unsynchronized audio and visual segments sampled from the same video, forcing the model to learn the relevant temporal features. At this time, not only the semantic correspondence was enforced between the video and the audio, but more importantly, the synchronization between them was also achieved.
    The researchers applied the curriculum learning strategy \cite{bengio2009curriculum} to this task and divided the samples into four categories: positives (the corresponding audio-video pairs), easy negatives (audio and video clips originating from different videos), difficult negatives (audio and video clips originating from the same video without overlap), and super-difficult negatives (audio and video clips that partly overlap), as shown in Fig. \ref{fig:representation-intro} (b).
	
		\begin{figure*}[t]
		\centering
		\includegraphics[width=0.8\linewidth]{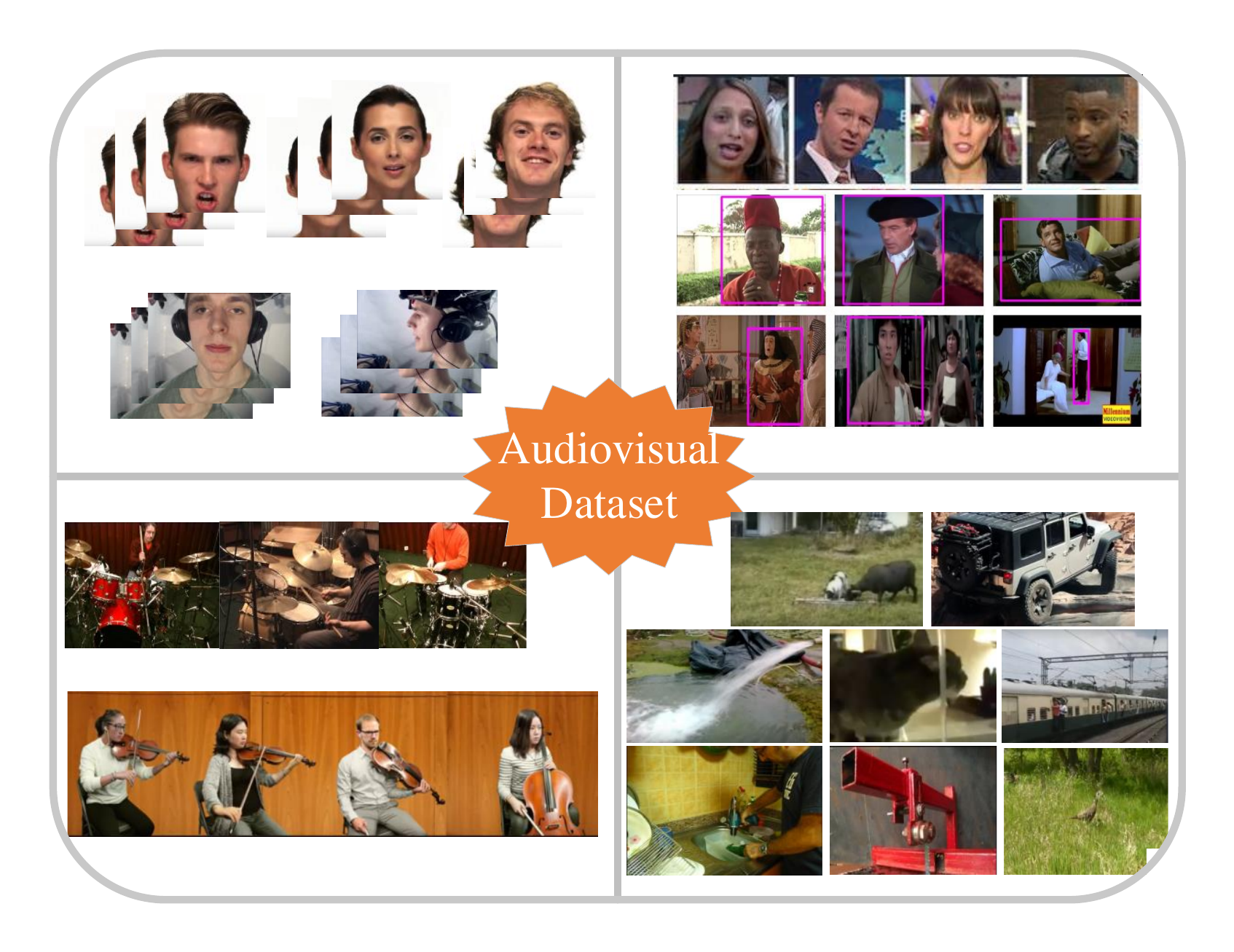}
		\caption{Demonstration of audio-visual datasets.}
		\label{fig:datasetvox}
	\end{figure*}
	
    The above studies rely on two latent assumptions: 1) the sound source should be present in the video, and 2) only one sound source is expected. However, these assumptions limit the applications of the respective approaches to real-life videos. 
    Therefore, Parekh et al. \cite{parekh2018weakly} leveraged class-agnostic proposals from both video frames to model the problem as a multiple-instance learning task for audio. As a result, the classification and localization problems could be solved simultaneously. The researchers focused on localizing salient audio and visual components using event classes in a weakly supervised manner. 
    This framework was able to deal with the difficult case of asynchronous audio-visual events.
    To leverage more detailed relations between modalities, Hu et al. \cite{hu2018deep} recommended a deep coclustering model that extracted a set of distinct components from each modality. The model continually learned the correspondence between such representations of different modalities. The authors further introduced K-means clustering to distinguish concrete objects or sounds. 
	

	\section{Recent Public Audio-visual Datasets}
	\label{sec:dataset}

	Many audio-visual datasets ranging from speech- to event-related data have been collected and released. 
	We divide datasets into two categories: audio-visual speech datasets that record human face with the corresponding speech, and audio-visual event datasets that consist of musical instrument videos and real events' videos. In this section, we summarize the information of recent audio-visual datasets (Table \ref{tbl:dataset_summary}). 
	
	\subsection{Audio-visual Speech Datasets}
	
	Constructing datasets containing audio-visual corpora is crucial to understanding audio-visual speech. The datasets are collected in lab-controlled environments where volunteers read the prepared phrases or sentences, or in-the-wild environments of TV interviews or talks. 
	
	\subsubsection{Lab-controlled Environment}
	\begin{table*}[t]
		\centering
		\caption{Summary of speech-related audio-visual datasets. These datasets can be used for all tasks related to speech we have mentioned above. Note that the length of a `speech' dataset denotes the number of video clips, while for `music' or 'real event' datasets, the length represents the total number of hours of the dataset.}
		\begin{tabular}{llllll}
			\hline
			\multicolumn{1}{l}{Category} & \multicolumn{1}{|l|}{Dataset} & \multicolumn{1}{l|}{Env.} & \multicolumn{1}{l|}{Classes} & \multicolumn{1}{l|}{Length*} & \multicolumn{1}{l}{Year} \\ \hline 
			
			& GRID \cite{cooke2006audio} &  Lab & 34 & 33,000 & 2006 \\
			& Lombard Grid \cite{alghamdi2018corpus} &  Lab & 54 & 54,000 &   2018 \\
			& TCD TIMIT \cite{harte2015tcd} & Lab  & 62 & - &   2015 \\
			& Vid TIMIT \cite{sanderson2009multi} & Lab  & 43 &-  &    2009 \\
			& RAVDESS \cite{livingstone2018ryerson} & Lab  & 24 &-  &  2018 \\
			& SEWA \cite{kossaifi2019sewa} &  Lab & 180 & - &  2017 \\
			Speech & OuluVS \cite{zhao2009lipreading} &  Lab & 20 & 1000 &  2009 \\
			& OuluVS2 \cite{anina2015ouluvs2} &  Lab & 52 & 3640 &  2016 \\
			& Voxceleb \cite{nagrani2017voxceleb} & Wild & 1,251    & 154,516 & 2017 \\
			& Voxceleb2 \cite{chung2018voxceleb2} & Wild & 6,112  & 1,128,246 & 2018 \\
			& LRW \cite{chung2016lip} & Wild & $\sim$1000 & 500,000 &  2016 \\
			& LRS \cite{chung2017lip}& Wild & $\sim$1000 & 118,116  &  2017 \\
			& LRS3 \cite{Chung17a} & Wild & $\sim$1000 & 74,564 & 2017 \\
			& AVA-ActiveSpeaker \cite{roth2019ava} & Wild & -  & 90,341 &  2019\\\hline
			& C4S \cite{bazzica2017vision} & Lab & - & 4.5 & 2017 \\
			Music & ENST-Drums \cite{gillet2006enst} & Lab & - & 3.75 & 2006 \\
			& URMP \cite{li2019creating} & Lab &  - & 1.3 &  2019 \\\hline
			& YouTube-8M \cite{abu2016youtube} & Wild & 3862 &  350,000 & 2016 \\
			& AudioSet \cite{gemmeke2017audio} & Wild & 632 &  4971 & 2016 \\
			Real Event& Kinetics-400 \cite{kay2017kinetics} & Wild & 400 & 850* &  2018 \\
			& Kinetics-600 \cite{carreira2018short} & Wild &  600 & 1400* &  2018 \\
			& Kinetics-700 \cite{carreira2019short} & Wild &  700 & 1806* &  2018 \\		& 
			
		\end{tabular}
		\label{tbl:dataset_summary}
	\end{table*}
	
	Lab-controlled speech datasets are captured in specific environments, where volunteers are required to read the given phases or sentences. Some of the datasets only contain videos of speakers that utter the given sentences; these datasets include GRID \cite{cooke2006audio}, TCD TIMIT \cite{harte2015tcd}, and VidTIMIT \cite{sanderson2009multi}. Such datasets can be used for lip reading, talking face generation, and speech reconstruction. Development of more advanced datasets has continued: e.g., Livingstone et al. offered the RAVDESS dataset \cite{livingstone2018ryerson} that contained emotional speeches and songs. The items in it are also rated according to emotional validity, intensity and authenticity.
	
	Some datasets such as Lombard Grid \cite{alghamdi2018corpus} and OuluVS \cite{zhao2009lipreading, anina2015ouluvs2} focus on multiview videos.
	In addition, a dataset named SEWA offers rich annotations, including answers to a questionnaire, facial landmarks, (low-level descriptors of) LLD features, hand gestures, head gestures, transcript, valence, arousal, liking or disliking, template behaviors, episodes of agreement or disagreement, and episodes of mimicry.

	\subsubsection{In-the-wild Environment}
	The above datasets were collected in lab environments; as a result, models trained on those datasets are difficult to apply in real-world scenarios. Thus, researchers have tried to collect real-world videos from TV interviews, talks and movies and released several real-world datasets, including LRW, LRW variants \cite{chung2016lip, chung2017lip, Chung17a}, Voxceleb and its variants \cite{nagrani2017voxceleb, chung2018voxceleb2}, AVA-ActiveSpeaker \cite{roth2019ava} and AVSpeech \cite{ephrat2018looking}. 
	The LRW dataset consists of 500 sentences \cite{chung2016lip}, while its variant contains 1000 sentences\cite{chung2017lip, Chung17a}, all of which were spoken by hundreds of different speakers. 
	VoxCeleb and its variants contain over 100,000 utterances of 1,251 celebrities \cite{nagrani2017voxceleb} and over a million utterances of 6,112 identities \cite{chung2018voxceleb2}, respectively. 
	
	AVA-ActiveSpeaker \cite{roth2019ava} and AVSpeech \cite{ephrat2018looking} datasets contain even more videos. 
	The AVA-ActiveSpeaker \cite{roth2019ava} dataset consists of 3.65 million human-labeled video frames (approximately 38.5 hrs)
	The AVSpeech \cite{ephrat2018looking} dataset contains approximately 4700 hours of video segments from a total of 290k YouTube videos spanning a wide variety of people, languages, and face poses. 
	The details are reported in Table \ref{tbl:dataset_summary}.

	\subsection{Audio-visual Event Datasets}
    Another audio-visual dataset category consists of music or real-world event videos. These datasets are different from the aforementioned audio-visual speech datasets in not being limited to facial videos.
    
    \subsubsection{Music-related Datasets}
    Most music-related datasets were constructed in the lab environment. For example, ENST-Drums \cite{gillet2006enst} merely contains drum videos of three professional drummers specializing in different music genres. The C4S dataset \cite{bazzica2017vision} consists of 54 videos of 9 distinct clarinetists, each performing 3 different classical music pieces twice 
(4.5h in total). 
    
    The URMP \cite{li2019creating} dataset contains a number of multi-instrument musical pieces. However, these videos were recorded separately and then combined. To simplify the use of the URMP dataset, Chen et al. further proposed the Sub-URMP \cite{chen2017deep} dataset that contains multiple video frames and audio files extracted from the URMP dataset.
    
    \subsubsection{Real Events-related Datasets}
    More and more real-world audio-visual event datasets have recently been released that consist of numerous videos uploaded to the Internet. The datasets often comprise hundreds or thousands of event classes and the corresponding videos. Representative datasets include the following.
    
    Kinetics-400 \cite{kay2017kinetics}, Kinetics-600 \cite{carreira2018short} and Kinetics-700 \cite{carreira2019short} contain 400, 600 and 700 human action classes with at least 400, 600, and 600 video clips for each action, respectively. Each clip lasts approximately 10 s and is taken from a distinct YouTube video. The actions cover a broad range of classes, including human-object interactions such as playing instruments, as well as human-human interactions such as shaking hands.  
    The AVA-Actions dataset \cite{gu2018ava} densely annotated 80 atomic visual actions in 43015 minutes of movie clips, where actions were localized in space and time, resulting in 1.58M action labels with multiple labels corresponding to a certain person.
    
    AudioSet \cite{gemmeke2017audio}, a more general dataset, consists of an expanding ontology of 632 audio event classes and a collection of 2,084,320 human-labeled 10-second sound clips. The clips were extracted from YouTube videos and cover a wide range of human and animal sounds, musical instruments and genres, and common everyday environmental sounds.
    YouTube-8M \cite{abu2016youtube} is a large-scale labeled video dataset that consists of millions of YouTube video IDs with high-quality machine-generated annotations from a diverse vocabulary of 3,800+ visual entities.
    
	\section{Discussion}
    Audio-visual learning (AVL) is a foundation of the multimodality problem that integrates the two most important perceptions of our daily life. Despite great efforts focused on AVL, there is still a long way to go for real-life applications. In this section, we briefly discuss the key challenges and the potential research directions in each category.
    
    \subsection{Challenges}
    The heterogeneous nature of the discrepancy in AVL determines its inherent challenges. 
    Audio tracks use a level of electrical voltage to represent analog signals, while the visual modality is usually represented in the RGB color space; the large gap between the two poses a major challenge to AVL. 
    The essence of this problem is to understand the relation between audio and vision, which also is the basic challenge of AVL. 
    
    \textbf{Audio-visual Separation and Localization} is a longstanding problem in many real-life applications. Regardless of the previous advances in speaker-related or recent object-related separation and localization, the main challenges are failing to distinguish the timbre of various objects and exploring ways of generating the sounds of different objects. Addressing these challenges requires us to carefully design the models or ideas (e.g., the attention mechanism) for dealing with different objects. \textbf{Audio-visual correspondence learning} has vast potential applications, such as those in criminal investigations, medical care, transportation, and other industries. Many studies have tried to map different modalities into the shared feature space. However, it is challenging to obtain satisfactory results since extracting clear and effective information from ambiguous input and target modalities remains difficult. Therefore, sufficient prior information (the specific patterns people usually focus on) has a significant impact on obtaining more accurate results. 
    \textbf{Audio and vision generation} focuses on empowered machine imagination. In contrast to the conventional discriminative problem, the task of cross-modality generation is to fit a mapping between probability distributions. Therefore, it is usually a many-to-many mapping problem that is difficult to learn. Moreover, despite the large difference between audio and visual modalities, humans are sensitive to the difference between real-world and generated results, and subtle artifacts can be easily noticed, which makes this task more challenging. 
    Finally, \textbf{audio-visual representation learning} can be regarded as a generalization of other tasks. As we discussed before, both audio represented by electrical voltage and vision represented by the RGB color space are designed to be perceived by humans while not making it easy for a machine to discover the common features. 
The difficulty stems from having only two modalities and lacking explicit constraints. Therefore, the main challenge of this task is to find a suitable constraint. Unsupervised learning as a prevalent approach to this task provides a well-designed solution, while not having external supervision makes it difficult to achieve our goal. The challenging of the weakly supervised approach is to find correct implicit supervision. 
    
    \subsection{Directions for Future Research}
    AVL has been an active research field for many years \cite{darrell2000audio, fisher2001audiovisualfusion} and is crucial to modern life. However, there are still many open questions in AVL due to the challenging nature of the domain itself and people's increasing demands. 
    
    First, from a macro perspective, as AVL is a classic multimodality problem, its primary issue is to learn the mapping between modalities, specifically to map the attributes in audio and the objects in an image or a video. We think that mimicking the human learning process, e.g., by following the ideas of the attention mechanism and a memory bank may improve performance of learning this mapping. Furthermore, the second most difficult goal is to learn logical reasoning. Endowing a machine with the ability to reason is not only important for AVL but also an open question for the entire AI community. Instead of directly empowering a machine with the full logic capability, which is a long way to go from the current state of development, we can simplify this problem and consider fully utilizing the prior information and constructing the knowledge graph. Building a comprehensive knowledge graph and leveraging it in specific areas properly may help machine thinking. 
    
    As to each task we have summarized before, Sec. \ref{sec:sep_loc} and Sec. \ref{sec:corr} can be referred to as the problem of `understanding', while Sec. \ref{sec:generation} and Sec. \ref{sec:representation_learning} can be referred to as `generation' and `representation learning' respectively. 
    Significant advances in understanding and generation tasks such as lip-reading, speaker separation, and talking face generation have recently been achieved for human faces. The domain of faces is comparatively simple yet important since the scenes are normally constrained, and it has a sizable amount of available useful prior information. For example, consider a 3d face model. These faces usually have neutral expressions, while the emotions that are the basis of the face have not been studied well. Furthermore, apart from faces, the more complicated in-the-wild scenes with more conditions are worth considering. 
    Adapting models to the new varieties of audio (stereoscopic audio) or vision (3D video and AR) also leads in a new direction. 
    The datasets, especially large and high-quality ones that can significantly improve the performance of machine learning, are fundamental to the research community \cite{sun2017revisiting}. However, collecting a dataset is labor- and time-intensive. Small-sample learning also benefits the application of AVL. Learning representations, which is a more general and basic form of other tasks, can also mitigate the dataset problem. While recent studies lacked sufficient prior information or supervision to guide the training procedure, exploring suitable prior information may allow models to learn better representations. 
    
    Finally, many studies focus on building more complex networks to improve performance, and the resulting networks generally entail unexplainable mechanisms. To make a model or an algorithm more robust and explainable, it is necessary to learn the essence of the earlier explainable algorithms to advance AVL. 
    
    \section{Conclusions}
    \label{sec:conclusion}
    The desire to better understand the world from the human perspective has drawn considerable attention to audio-visual learning in the deep learning community. This paper provides a comprehensive review of recent advances in audio-visual learning categorized into four research areas: audio-visual separation and localization, audio-visual correspondence learning, audio and visual generation, and audio-visual representation learning. Furthermore, we present a summary of datasets commonly used in audio-visual learning. The discussion section identifies the key challenges of each category followed by potential research directions.

	\bibliographystyle{IEEEtran}
\bibliography{avc-review}
		
\end{document}